\documentclass[titlepage]{book}

\usepackage{graphicx, amssymb, amsmath, enumerate, mathrsfs, dsfont}
\usepackage{kotex}
\usepackage{epsfig, color, tcolorbox}
\usepackage[utf8]{inputenc}
\usepackage{caption}
\usepackage{subcaption}
\usepackage{makeidx, minitoc}
\usepackage{fancyhdr}
\usepackage[colorlinks=false]{hyperref}
\hypersetup{pdfborder=0 0 0}
\usepackage{cancel}

\usepackage{natbib}

\usepackage{paralist}

\graphicspath{ {./images/} }

\usepackage{tikz}
\usetikzlibrary{bayesnet}

\newcommand{\indep}[0]{\ensuremath{\perp\!\!\!\perp}}

\title{A Brief Introduction to \\ Causal Inference in Machine Learning}

\author{
    Kyunghyun Cho \\
    New York University \& Genentech
}

\date{2024}

\begin{document}

% TITLE

\maketitle

% % PREFACE

\begin{center}
    \textbf{\Large{PREFACE}}
\end{center}

This is a lecture note produced for DS-GA 3001.003 ``Special Topics in DS - Causal Inference in Machine Learning'' at the Center for Data Science, New York University in Spring, 2024. This course was created to target master's and PhD level students with basic background in machine learning but who were not exposed to causal inference or causal reasoning in general previously. In particular, this course focuses on introducing such students to expand their view and knowledge of machine learning to incorporate causal reasoning, as this aspect is at the core of so-called out-of-distribution generalization (or lack thereof.) 

This lecture note does not follow a traditional curriculum for teaching causal inference. This lecture note does not subscribe solely to either the potential outcome framework or the do-calculus framework, but is rather flexible in taking concepts and ideas from these two camps (which after all do look more or less the same) in order to build up the foundation of causal inference from the first principles. In doing so, the first half of this note covers a variety of basic topics, including probabilistic graphical models, structural causal models, causal quantities of interest, conditional vs. interventional probabilities, regression, randomized controlled trials, bandit algorithms, inverse probability weighting, matching and instrumental variables. I do not go too deep into each of these topics, although the emphasis is given to how these topics are all connected with each other (and sometimes are equivalent.) For this first half of the course, I read and consulted the following books (lightly only, though) and recommend students go deeper into these books if they are interested in learning more about causal inference:

\vspace{3mm}
\begin{compactenum}
    \item Pearl. Causality. 2nd eds. 2009. \citep{pearl2009causality}
    \item Imbens \& Rubin. Causal inference in statistics, social, and biomedical sciences. 2015. \citep{imbens2015causal}
    \item Cunningham. Causal Inference: the Mixtape. 2021. \citep{cunningham2021causal}
\end{compactenum}
\vspace{3mm}

Based on the foundation built in the first half (or more like two thirds) of the course, the course takes a turn toward generalization in machine learning. In particular, I try to argue that the probabilistic graphical model based framework from causal inference can be an invaluable tool for specifying and understanding so-called out-of-distribution generalization. I draw (coarse) connections from causal inference to the following ideas in machine learning, to demonstrate this point:

\vspace{3mm}
\begin{compactenum}
    \item Distributional shifts
    \item The principle of invariance
    \item Preference-based learning for language models
\end{compactenum}
\vspace{3mm}

To be very honest, this is a very thin lecture note for a course with a very thin content. This note should be considered as the very first sign post at the entrance to a huge forest called causality, and nothing more. If you want to expand slightly a bit more, see this short introductory material I have written together with my PhD student, Jiwoong Daniel Im~\citep{im2023active}. 

Finally, I am infinitely grateful to Daniel Im, Divyam Madaan and Taro Makino for helping me as amazing teaching assistants in preparing the lecture note as well as giving the lab sessions in Spring 2024. The lab materials they have prepared are all available at \url{https://github.com/kyunghyuncho/2024-causal-inference-machine-learning}.

% \firstword{T}{his} is a preface! \lipsum[1]

% \vspace{\baselineskip}

% \textit{Some Left Text, (Maybe a Date)} \hfill PREFACE AUTHOR

% TABLE OF CONTENTS

\tableofcontents
\clearpage

% CONTENTS

\pagenumbering{arabic}

\chapter{Probabilistic Graphical Models}

The goal of causal inference is to figure out a causal relationship, or the lack thereof, between two sets of variables; causes and outcomes. It is thus only natural to think of how we determine whether any particular variable is a cause or an outcome. It is often relatively more straightforward to determine what an outcome variable is, as this determination is done based on our subjective interest. For instance, a typical outcome variable in medicine is a disease-free survival rate within some years since diagnosis or treatment. This choice is natural, since this variable, or its quantity, is what we want to maximize. It is however much less clear how to determine which variable should be considered a cause. For instance, in the classical example of `smoking causes lung cancer', what makes us choose `whether someone smokes ciagarettes' as a cause variable rather than `a mutation in a particular gene'? It becomse even more mind-boggling once we realize that this choice of `smoking' as a cause meant that we decided to ignore many variables, such as `whether a farmer decided to grow tobacco'. 

It is thus an important, if not the most important, job for practitioners of causal inference to convincingly argue why some variables are included and others were omitted. They also must argue why some of the included variables are considered potential causes and why they chose a particular variable as an outcome. This process can be thought of as defining a small universe in which causal inference must be performed. There are many different ways to define and describe such a universe, and in this lecture note, we largely stick to using a probabilistic graphical model, or a corresponding structural causal model, as a way to describe each universe, which is the main topic of this chapter.

\section{Probababilistic Graphical Models}
\label{sec:pgm}

In this course, we rely on probabilistic graphical models quite extensively in order to describe their statistical and causal relationships among random variables. A probabilistic graphical model, which is also referred as a Bayesian graphical model, is a directed graph $G=(V,E)$, where $V$ is a set of vertices/nodes and $E$ is a set of directed edges. Each node $v \in V$ corresponds to a random variable, and each edge $e = (v_s, v_e)$ represents the dependence of $v_e$ on $v_s$. Let us for now assume that this graph is acyclic, that is, there is no cycle within this graph. In other words, $G$ is a directed acyclic graph, throughout the rest of this course. 

For each node $v \in V$, we define a probability distribution $p_v(v | \mathrm{pa}(v))$ over this variable conditioned on all the parent nodes 
\begin{align}
    \mathrm{pa}(v) = \left\{ v' \in V | 
    (v', v) \in E
    \right\}.
\end{align}
We can then write a joint distribution over all the variables as
\begin{align}
    p_V (V) = \prod_{v \in V} p_v(v | \mathrm{pa}(v)),
\end{align}
following the chain rule of probabilities. 
When $\mathrm{pa}(v) = \emptyset$, i.e. $v$ does not have any incoming edges, we call the associated distribution $p_v(v)$ a prior distribution. 

With $P$ a set of all conditional probabilities $p_v$'s, we can denote any probabilistic graphical model as a triplet $(V, E, P)$. 

From this joint distribution, we can derive all kinds of conditional distributions by marginalizing variables and applying the definition of a conditional probability. If we are not interested in a particular node $\bar{v} \in V$, we can marginalize out this variable by
\begin{align}
    p(V \backslash \{\bar{v}\}) = \sum_{\bar{v}} p_V (V).
\end{align}
If $\bar{v}$ is a continuous random variable, we replace $\sum$ with $\int$. We can always turn a joint probability into a conditional probability by
\begin{align}
    p(V \backslash \{\tilde{v}\} | \tilde{v}) 
    =
    \frac{p_V(V)}{p_{\tilde{v}}(\tilde{v})}.
\end{align}

Using the definition of the conditional probability, we can write marginalization in the following way:
\begin{align}
    p(V \backslash \{\bar{v}\}) = \sum_{\bar{v}} \frac{p_V(V)}{p_{\tilde{v}}(\tilde{v})} p_{\tilde{v}}(\tilde{v})
    = \sum_{\bar{v}} p(V \backslash \{\bar{v}\}) p_{\tilde{v}}(\tilde{v}).
\end{align}
Marginalization corresponds to computing the weighted sum of the conditional probability of the remaining variables according to the marginal probability of the variable being marginalized. 

Let's assume we can sample readily from any conditional probability $p_v$ with $v\in V$. We can then draw a sample from the joint distribution readily by breadth-first sweeping of all the variables. That is,
\begin{align}
    \tilde{v} \sim p_v (v | \tilde{\mathrm{pa}}(v)),
\end{align}
where $\tilde{\mathrm{pa}}(v) = \left\{ \tilde{v}' \sim p_{v'}(v' | \tilde{\mathrm{pa}}(v')) | v' \in \mathrm{pa}(v) \right\}$. This procedure is called ancetral sampling and is an exact, unbiased way to sample from this joint distribution. 

If we set aside efficiency, ancestral sampling is an extremely powerful tool, as it allows us to sample from any marginal distribution as well as any conditional distribution. In the former case, we simply discard the draws that correspond to the uninteresting variables (that are being marginalized). In the latter case, we only keep samples whose values, corresponding to the conditioning variables (that are on the right hand side of $|$), are precisely those that we want the distribution to be conditioned on. Both of these approaches are not efficient, and it is often much better to use a more sophisticated sampling algorithm, often based on the idea of Markov Chain Monte Carlos (MCMC). 

Any probability distribution can be expressed as a table (though, this table may have infinitely many rows) that consists of two columns. The first column takes the value of interest and the second column its probability (either density or mass). The probability function $p_v$ above works by hashing $v$ into the row index in this table and retrieving the associated probability, i.e. $p_v: \mathbb{V} \to \mathbb{R}_+$. This function satisfies the following normalization property:
\begin{align}
    1 =
    \begin{cases}
        \sum_{v \in \mathbb{V}} p_v (v),&\text{ if } v \text{ is discrete.} \\
        \int_{v \in \mathbb{V}} p_v (v) \mathrm{d}v,&\text{ if } v \text{ is continuous.} \\
    \end{cases}
\end{align}

This view allows us to effectively turn a set of samples into the original distribution from which these samples were drawn (of course, with some variance.) Let $S = \left( \tilde{v}_1, \tilde{v}_2, \ldots, \tilde{v}_N \right)$ be a multi-set of samples drawn from an unknown distribution over a discrete variable $v$, without loss of generality. We can then recover the original distribution by constructing a table where each row is
\begin{align}
    \left(v, \left. \sum_{n=1}^N \mathds{1}(\tilde{v}_n = v) \right/ N \right),
\end{align}
with $v \in \mathbb{V}$. This corresponds to maximum likelihood learning without any regularization.

In this table, we can think of all these rows' contents as the parameters of this model we have built to approximate the underlying distribution from which $S$ was drawn. This view will come handy later when we discuss using a deep neural network instead of an explicit table to represent a probability distribution. 

A major downside of this explicit table based approach is that $q_v(v) = 0$ for all $v \not \in S$. Such an extreme value (the probability of $0$) should not be used when we estimate these probabilities from a small number of data points, since we cannot rule out the fact that we simply did not see that particular instance just because we did not draw enough samples. Furthermore, this prevents us from properly defining a conditional probability $p_{v'} (v' | v)$, since
\begin{align}
    p_{v'} (v' | v) = \frac{p_{v',v} (v', v)}{p_v(v)}.
\end{align}
If $v$ is set such that $p(v) = 0$, this conditional probability is not well defined. We thus have to regularize maximum likelihood learning. 

This probabilistic graphical model is a good way to abstract out some of the details on how to implement individual probability distributions for studying causal inference in machine learning, as it frees us from worrying about the aspect of learning until it is absolutely necessary. Learning in this context refers to inferring the underlying distribution from which data points were drawn, and the table-based approach above is the most naive one that is not really useful in practice. We can however for now assume that this table-based approach works well and continue studying causal inference with already inferred conditional distributions. 

\section{Structural Causal Models}
\label{sec:scm}

Although a directed edge in the probabilistic graphical model looks like it encodes the order in which the variables are generated, it is not necessarily so from the perspective of probability. According to the Bayes' rule,
\begin{align}
    p_v(v | v') = \frac{p_{v'}(v' | v) p_v(v)}{p_{v'}(v')}.
\end{align}
This implies that we can always flip the arrow of the edge between $v$ and $v'$ without altering the joint as well as conditional probabilities.\footnote{
    We will assume from here on that $p(x) > 0$ for any $x$ and $p$.
}

This lack of direct relevance of the direction of each edge in $G$ to the joint probability raises a lot of confusion, when we try to use the probabilistic graphical model in the context of causal inference and reasoning. Instead, we may want to use a slightly different way to express the same generative process underlying a given probability graphical model $G=(V,E)$. 

We do so by writing the process of sampling a value associated with each variable $v \in V$ rather than its distribution, as the combination of a deterministic function $f_v$ and external (exogenous) noise $\epsilon_v$:
\begin{align}
    v \leftarrow f_{v}(\mathrm{pa}(v), \epsilon_v).
\end{align}
This says that the value of $v$ is computed based on the values of its parent nodes $\mathrm{pa}(v)$ and external noise $\epsilon_v$ by the deterministic function $f_{v}$. 

This way is much more explicit about the generating process than before, since the use of the function $f$ clearly suggests that perturbing the output of the function would not change the input to the function, although perturbing the input to the function would change the output. For instance, you can imagine that $f_v$ corresponds to pushing a book on your desk using your hand with force of $v'$ and that $v$ encodes the new position of the book. $\epsilon_v$ can be an unknown level of friction cause by your earlier (but forgotten) choice of your desk. Changing force $v'$ of course affects the new position of the book $v$ together with some changing level of $\epsilon_v$, but changing the position of the book would not change the force I have not applied yet to the book. 

We call this representation of the generative process a structural causal model. Similarly to the probabilistic graphical model above, we can represent any structural causal model as a triplet $(V, F, U)$, where $V$ is a set of variables, $F$ is a set of corresponding functions and $U$ is a set of corresponding noise variables. 

Any structural causal model can be turned into a probabilistic graphical model by using the change of variables, i.e., $f_{v}(\mathrm{pa}(v), \cdot)$ and assuming the known prior distribution over the external noise variable $\epsilon_v$. Or, more simply, we can do so because we can find a distribution over $v \sim f_v (\mathrm{pa}(v), \epsilon_v)$.

We can draw samples from any given structural causal model, just like with the probabilistic graphical models above, by ancestral sampling. Once we have samples from the joint distribution, we can infer various conditional distributions. Among these, the posterior distribution over the external noise variables is of a particular interest for the purpose of counterfactual reasoning. Let $q(U)$ be a distribution that corresponds to all samples of $\epsilon_v$'s that led to a particular configuration $\hat{V}$. Then the posterior distribution $q(v)$ can be thought of as the distribution over the external (uncontrollable) factors that led to the particular outcome. We then can answer the question what would have happened to a target variable $v$ had some of the variables were set differently, by fixing the external factors to follow $q(v)$ and the rest of the variables to the original values $\hat{V}$. This corresponds to counterfactual reasoning (had I done something differently, what would have happened?)

\section{Learning and a generative process}
\label{sec:learning}

Learning, in machine learning, refers to the process by which we configure a black box predictive model to capture the predictive relationship between a set of variables. In the simplest form, let us consider having two variables; input $v$ and output $v'$. $g_\theta$ is a predictive function that models the relationship between $v$ and $v'$ by mapping an instance of $v$ to the corresponding $v'$, i.e., $g_{\theta}(v')$ is the prediction of $v'$ given $v$. $\theta$ is a collection of parameters that a learning algorithm configures to make $g$ as predictive of $v$ given $v'$ as possible. 

Learning starts from data which is a set of examples drawn from an unknown data generating process $\mathcal{G}$.
This data generating process can be described using either a probabilistic graphical model or equivalently a structural causal model. 
Let $D=\left\{ (v_1, v_1'), \ldots, (v_N, v_N') \right\}$ be our training dataset. The goal is then to use this dataset to infer $\theta$ that would make $g$ highly predictive of $v$ given $v'$. 

There are many ways to measure how predictive $g$ is given a pair $(v, v')$, but we use here a log probability:
\begin{align}
    r(\theta; v,v') = \log p(v | v'; g_\theta(v')).
\end{align}
This means that $g_{\theta}(v')$ parameterized a conditional distribution over $v$ given $v'$, or equivalently, $g_{\theta}(v')$ outputs a conditional distribution over $v$. If this quantity is large, it means that $v$ is highly probable under the predictive distribution by $g$ given $v'$. 

Learning is then to solve the following optimization problem:
\begin{align}
    \arg\max_{\theta} \frac{1}{N} \sum_{n=1}^N r(\theta; v_n, v_n').
\end{align}
Once learning is over, we can readily compute
\begin{align}
    p (v|v') \approx p(v|v'; g_{\theta}(v')) = p(v|v'; \theta).
\end{align}

If we assume that $p(v | v'; \theta)$ is a great approximation of $p(v|v')$, we can use the former in place of the latter without too much worry. In other words, learning corresponds to figuring out a conditional distribution $p(v|v')$ from data. This data was produced from the underlying data generating process $\mathcal{G}$ which may have more variables in addition to $v$ and $v'$.

From this description of learning, it is immediately clear how this can be a replacement of the table-based approach from earlier, and that the table-based approach earlier was a special case of this learning-based approach, where $\theta$ corresponded to all the entries within a table. Once we take this learning-based approach, we can free ourselves from having to explicitly construct a large table and can also use various regularization techniques to avoid the issue of $0$ probability as well as benefit from generalization.

Let $\mathcal{G}=(V, E, P)$ with $v, v' \in V$ and $\overline{V} = V \backslash \left\{ v, v' \right\}$. Then,
\begin{align}
    p(v | v') = \frac{\sum_{\overline{V}} p(\{v,v'\} \cup \overline{V})}{\sum_{\{v\} \cup \overline{V}} p(\{v'\} \cup V')}.
\end{align}
That is, we marginalize out all variables in $\overline{V}$ and then divide it by the marginal probability of $v'$ to get the conditional probability of $v$ given $v'$. 

On one hand, this tells us that learning allows us to recover any arbitrary conditional distribution induced by an underlying data generating process as long as we have data produced following the same data generating process. On the other hand, this also tells us that there can be many different data generating processes that result in the exactly same conditional distribution given a pair of variables. In fact, as long as we do not use all variables from the generating process, there will always be ambiguities that cannot be resolved based on the learning procedure laid out above. 

As an example, consider the following two structural causal models:

\paragraph{Causal Model 1.}
\begin{align}
    &v' \leftarrow \epsilon_{v'} \\
    &v^l \leftarrow v' + a + \epsilon_{v^l} \\
    &v^r \leftarrow v' + b + \epsilon_{v^r} \\
    &v \leftarrow v^l + v^r + \epsilon_{v},
\end{align}
where $\epsilon_{v^l}$ and $\epsilon_{v^r}$ both follow standard Normal distributions ($\mathcal{N}(0, 1^2)$.
$\epsilon_{v}$ also follows standard Normal distribution.

\paragraph{Causal Model 2.}
\begin{align}
    &v' \leftarrow \epsilon_{v'} \\
    &v^c \leftarrow v' + a + b + \epsilon_{v^c} \\
    &v \leftarrow v^c + \epsilon_{v},
\end{align}
where $\epsilon_{v^c} \sim \mathcal{N}(0, 2)$. 

Then, $p(v|v') = \mathcal{N}(v; v'+a+b, 3)$ for both causal models, although these two models are very different. 

This ambiguity plays an important role in both causal inference and so-called out-of-distribution generalization. We will study both more carefully later in the course.

\section{Latent variable models are not necessarily causal models}

When we build a predictive model on a pair $(v, v')$, there are variables that are left out from the original data generating process. Those unused variables may be the ones for which we simply threw away observations, because they were not in our interest, or the ones we actually do not observe. These unobserved variables contribute to the complexity of $p(v|v')$ via the process of marginalization.

It is not trivial to build a predictive model, or equivalently in our context a deep neural network, that outputs a highly complex predictive distribution, as these complex distributions often do not have easily-parametrizable analytical forms. In these cases, there are two commonly used approaches; (1) autoregressive models and (2) latent variable models. The former relies on the chain rule of probabilities and decomposes the conditional probability as a product of coordinate-wise conditional probabilities:
\begin{align}
    p(v|v'; \theta) = \prod_{i=1}^d p(v_i | v'; \theta),
\end{align}
where $v = [v_1, \ldots, v_d]$. 
By assuming that each coordinate-wise conditional probability is of a simpler form, we can still use a simple deep neural network to approximate a highly complex joint conditional probability. Because each conditional probability on the right hand side is parametrized with the same set of parameters $\theta$, it is possible to build such an autoregressive model for a variable-sized observation, that is, $\mathrm{dim}(v)$ is not fixed \textit{a priori}. This approach is behind the recently successful and popular large-scale language models~\citep{brown2020language}.

Unlike autoregressive modeling, latent variable models explicitly introduce an unobserved variable $u$ that represents the missing portion of the underlying data generating process. Just like the missing (unobserved) portion gave rise to the highly complex predictive distribution via the process of marginalization, the introduced (anonymous) latent variable $u$ does the same:
\begin{align}
    p(v|v'; \theta) = \sum_{u} p_u(u) p(v|v', u; \theta).
\end{align}
If $u$ is continuous, we replace $\sum$ with $\int$. 

Because this marginalization is difficult almost always, it is natural to resort to sampling-based approximation. Because we are often interested in gradient-based learning, we consider sampling-based gradient approximation:
\begin{align}
    \nabla \log p(v|v'; \theta) 
    &=
    \sum_u
    \frac{p_u (u) p(v|v', u; \theta) \nabla \log p(v|v', u; \theta) }
    {\sum_{u'} p_u(u') p(v|v', u'; \theta)}
    \\
    &=
    \sum_u 
    p(u | v, v'; \theta)
    \nabla \log p(v|v', u; \theta)
    \\
    &\approx
    \frac{1}{M}
    \sum_{m=1}^M
    \nabla \log p(v|v', u^m; \theta),
\end{align}
where $u^m$ is the $m$-th posterior sample. 

It is however as challenging to compute the posterior distribution $p(u | v, v'; \theta)$ nor to sample from it. It is thus more common these days to maximize the lower bound to $p(v|v'; \theta)$ while amortizing approximate posterior inference into a separate neural network. This is called a variational autoencoder~\citep{kingma2013auto}. 
% We will study the variational autoencoder and its implication in causal inference for machine learning later in the semester.

Despite the seemingly similarity, these latent variables are not closely related to actual variables in the data generating process. They may or may not. If they indeed correspond to actual variables in the data generating process that we simply did not observe nor decided not to use data from, we may be able to derive conditions under which we can identify these unobserved variables and their associated distributions from partial data alone. It is however more common and realistic to think of these latent variables as a means to improving the expressive power of a deep neural network. In other words, latent variables are useful even if there are truly two variables, $v$ and $v'$, in the data generating process, since true $p_v(v|v')$ may be complicated on its own. Nevertheless, it is a useful tool to model any complex distribution, and thus we have spent a little bit of time discussing them.

\section{Summary}

In this chapter, we have established the very foundations on which we can discuss causal inference and its relationships to machine learning in the rest of the course:

\begin{enumerate}
    \item A brief discussion on the necessity of defining a universe over which causal inference is performed;
    \item Two (equivalent) ways to define a universe, probabilistic graphical models and structural causal models;
    \item What learning is, once the universe is defined.
\end{enumerate}

Based on these, we begin our journey into causal inference in machine learning.

\chapter{A Basic Setup}

\section{Correlation, Independence and Causation}

Two random variables, $u$ and $v$, are independent if and only if
\begin{align}
    p(u, v) = p(u) p(v).
\end{align}
That is, the probability of $u$ taking a certain value is not affected by that of $v$ taking another value. We say $u$ and $v$ are dependent upon each other when the condition above does not hold. 

In everyday life, we often confuse dependence with correlation, where two variables are correlated if and only if
\begin{align}
    \mathrm{cov}(u,v) = \mathbb{E}\left[ (u - \mu_u) (v - \mu_v) \right] > 0,
\end{align}
where $\mu_u = \mathbb{E}[u]$ and $\mu_v = \mathbb{E}[v]$. When this covariance is $0$, we say these two variables are uncorrelated. 

Despite our everyday confusion, these two quantities are only related and not equivalent. 
When two variables are independent, they are also uncorrelated, but when two variables are uncorrelated, they may not be independent. Furthermore, it turned out these two quantities are also only remotely related to the existence/lack of causation between two variables. 

You must have heard of the statement ``correlation does not imply causation.'' There are two sides to this statement. First, the existence of correlation between two variables does not imply that there exists a causal relationship between these two variables. An extreme example of this is tautology; if the relationship between $u$ and $v$ is identity, there is no causal relationship but the correlation between these two is maximal. 

Second, the lack of correlation does not imply the lack of causation. This is the more important aspect of this statement. Even if there is no correlation between two variables, there could be a causal mechanism behind these two variables. Although it is a degenerate case, consider the following structural causal model:
\begin{align}
    &a \leftarrow -u + \epsilon_a \\
    &b \leftarrow +u + \epsilon_b \\
    &v \leftarrow a + b + \epsilon_v.
\end{align}
The value of $v$ is caused by $u$ via two paths; $u \to a \to v$ and $u \to b \to v$, but these paths cancel each other. If we only observe $(u,v)$ pairs, it is easy to see that they are uncorrelated, since $v$ is constant. We will have more discussion later in the semester, but it is good time for you to pause and think whether these two paths matter, since they cancel each other. 

Consider as another example the following structural causal model:
% https://stats.stackexchange.com/a/357275
\begin{align}
    &z \leftarrow \epsilon_z \\
    &u \leftarrow 0.2 z + \sqrt{1.04} \epsilon_u \\
    &v \leftarrow 0.1 u - 0.5 z + 0.1 \epsilon_v,
\end{align}
where $\epsilon_z$, $\epsilon_u$ and $\epsilon_v$ are all standard Normal variables. Again, the structural causal model clearly indicates that $u$ causally affects $v$ via $0.1 u$, but the correlation between $u$ and $v$ is $0$ when we consider those two variables alone (that is, after marginalizing out $z$). 

This second observation applies equally to independence. That is, the independence between two variables does not imply the lack of a causal mechanism between two variables. The examples above apply here equally, as two uncorrelated Normal variables are also independent. 

This observation connects to an earlier observation that there are potentially many data generating processes that give rise to the same conditional distribution between two sets of variables. Here as well, the independence or correlatedness of two variables may map to many different generating processes that encode different causal mechanisms behind these variables. In other words, we cannot determine the causal relationship between two variables (without loss of generality) without predefining the underlying generating process (in terms of either the probabilistic graphical model or equivalently the structural causal model.)\footnote{
    There are algorithms to discover an underlying structural causal model from data, but these algorithms also require some assumptions such as the definition of the goodness of a structural causal model. This is necessary, since these algorithms all work by effectively enumerating all structural causal models that can produce data faithfully and choosing the best one among these. 
}

In other words, we must consider both variables of interest and the associated data generating model in order to determine whether there exists a causal relationship between these variables and what that relationship is. 

\section{Confounders, Colliders and Mediators}
\label{sec:confounders-colliders-mediators}

Let us consider a simple scenario where there are only three variables; $u$, $v$ and $w$. We are primarily interested in the relationship between the first two variables; $u$ and $v$. We will consider various ways in which these three variables are connected with each other and how such wiring affects the relationship between $u$ and $v$. 

\paragraph{Directly connected.}

Consider the following probabilistic graphical model. 

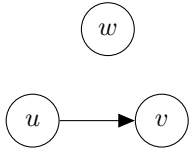
\begin{center}
\vspace{5mm}
\begin{tikzpicture}
  % Define nodes
  \node[latent] (u) {$u$}; % u node
  \node[latent, right=1cm of u] (v) {$v$}; % v node
  \node[latent, above=0.5cm of u, xshift=1cm] (w) {$w$}; % w node
  
  % Connect the nodes
  \edge{u}{v}; % Edge from u to v 
\end{tikzpicture}
\vspace{5mm}
\end{center}

$w$ does not affect either $u$ nor $v$, while $u$ directly causes $v$. In this case, the causal relationship between $u$ and $v$ is clear. If we perturb $u$, it will affect $v$, according to the conditional distribution $p_v(v | u)$. This tells us that we can ignore any node in a probabilistic graphical model that is not connected to any variable of interest. 

\paragraph{An observed confounder.}

Consider the following probabilistic graphical model, where $w$ is shaded, which indicates that $w$ is observed. 

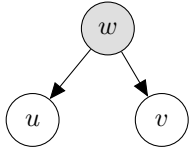
\begin{center}
\vspace{5mm}
\begin{tikzpicture}
  % Define nodes
  \node[latent] (u) {$u$}; % u node
  \node[latent, right=1cm of u] (v) {$v$}; % v node
  \node[obs, above=0.5cm of u, xshift=1cm] (w) {$w$}; % w node
  
  % Connect the nodes
  \edge{w}{u}; 
  \edge{w}{v}; 
\end{tikzpicture}
\vspace{5mm}
\end{center}

In this graph, the value/distribution of $u$ and that of $v$ are both determined individually already because we have observed $w$. This corresponds to the definition of conditional independence; 
\begin{align}
    p(u,v |w) = p(u|w) p(v|w).
\end{align}
Because the edge is directed from $w$ to $u$, perturbing $u$ does not change the observed value of $w$. The same applies to $v$ as well, since perturbing $v$ does not affect $u$, since this path between $u$ and $v$ via $w$ is blocked by observing $w$. 

\paragraph{An unobserved confounder.}

Consider the case where $w$ was not observed.

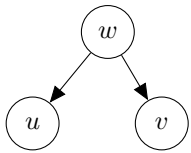
\begin{center}
\vspace{5mm}
\begin{tikzpicture}
  % Define nodes
  \node[latent] (u) {$u$}; % u node
  \node[latent, right=1cm of u] (v) {$v$}; % v node
  \node[latent, above=0.5cm of u, xshift=1cm] (w) {$w$}; % w node
  
  % Connect the nodes
  \edge{w}{u}; 
  \edge{w}{v}; 
\end{tikzpicture}
\vspace{5mm}
\end{center}

We first notice that in general 
\begin{align}
    p(u,v) = \int_w p(u|w)p(v|w)p(w) \mathrm{d}w \neq q(u)q(v),
\end{align} 
implying that $u$ and $v$ are not independent, unlike when $w$ was not observed. In other words, $u$ and $v$ are not conditionally independent given $w$. 
Perturbing $u$ however still does not affect $v$, since the value of $w$ is not determined by the value of $u$ according to the corresponding causal structural model. That is, $u$ does not affect $v$ causally (and vice versa.) This is the case where the independence and causality start to deviate from each other; $u$ and $v$ are not independent but each is not the cause of the other. Analogous to the former case of the observed $w$, where we say the path $u \leftarrow w \to v$ was closed, we say that the same path is open in this latter case. 

Because of this effect $w$ has, we call it a confounder. The existence of a confounder $w$ makes it difficult to tell whether the dependence between two variables we see is due to a causal relationship between these variables. $w$ confounds this analysis.

% {\bf TODO}: An example - comorbidity as $w$, a new treatment as $u$ and the output as $v$. even if $u$ doesn't affect $v$, because of $w$, we may falsely conclude that $u$ negatively affects $v$.

\paragraph{An observed collider.}

Consider the following graph where the arrows are flipped.

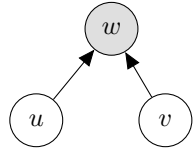
\begin{center}
\vspace{5mm}
\begin{tikzpicture}
  % Define nodes
  \node[latent] (u) {$u$}; % u node
  \node[latent, right=1cm of u] (v) {$v$}; % v node
  \node[obs, above=0.5cm of u, xshift=1cm] (w) {$w$}; % w node
  
  % Connect the nodes
  \edge{u}{w}; 
  \edge{v}{w}; 
\end{tikzpicture}
\vspace{5mm}
\end{center}

In general,
\begin{align}
    p(u,v | w) = 
    \frac{p(w|u, v) p(u) p(v)}
    {\int \int p(w|u, v) p(u) p(v) \mathrm{d}u \mathrm{d}v}
    \neq 
    q(u) q(v),
\end{align}
which means that $u$ and $v$ are not independent conditioned on $w$. 
This is sometimes called the explaining-away effect, because observing $w$ explains away one of two potential causes behind $w$. 

Although $u$ and $v$ are not independent in this case, there is no causal relationship between $u$ and $v$. $w$ is where the causal effects of $u$ and $v$ collide with each other (hence, $w$ is a collider) and does not pass along the causal effect between $u$ and $v$. Similarly to the case of an unobserved confounder above, this is one of those cases where independence does not imply causation.

We say that the path $u \to w \leftarrow v$ is open.

\paragraph{An unobserved collider.}

Consider the case where the collider $w$ is not observed. 

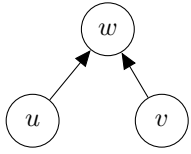
\begin{center}
\vspace{5mm}
\begin{tikzpicture}
  % Define nodes
  \node[latent] (u) {$u$}; % u node
  \node[latent, right=1cm of u] (v) {$v$}; % v node
  \node[latent, above=0.5cm of u, xshift=1cm] (w) {$w$}; % w node
  
  % Connect the nodes
  \edge{u}{w}; 
  \edge{v}{w}; 
\end{tikzpicture}
\vspace{5mm}
\end{center}

By construction, $u$ and $v$ are independent, as
\begin{align}
    p(u,v) = \int p(u) p(v) p(w|u,v) \mathrm{d}w 
    = p(u) p(v) \int p(w | u, v) \mathrm{d}w
    = p(u) p(v).
\end{align}
Just like before, neither $u$ or $v$ is the cause of the other. This path is closed. 

\paragraph{An observed mediator.}

Consider the case where there is an intermediate variable between $u$ and $v$:

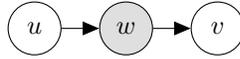
\begin{center}
\vspace{5mm}
\begin{tikzpicture}
  % Define nodes
  \node[latent] (u) {$u$}; % u node
  \node[obs, right=0.5cm of u] (w) {$w$}; % w node
  \node[latent, right=0.5cm of w] (v) {$v$}; % v node
  
  % Connect the nodes
  \edge{u}{w}; 
  \edge{w}{v}; 
\end{tikzpicture}
\vspace{5mm}
\end{center}

Because
\begin{align}
    p(u,v | w) = p(u) p(w|u) p(v|w)
    = q(u|w) q(v|w)
\end{align}
$u$ and $v$ are independent conditioned on $w$. 
However, perturbing $u$ does not affect $v$, since the value of $w$ is observed (that is, fixed.) 
We say that $u \to w \to v$ is closed in this case, and independence implies the lack of causality. 

\paragraph{An unobserved mediator.}

What if $w$ is not observed, as below?

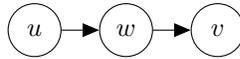
\begin{center}
\vspace{5mm}
\begin{tikzpicture}
  % Define nodes
  \node[latent] (u) {$u$}; % u node
  \node[latent, right=0.5cm of u] (w) {$w$}; % w node
  \node[latent, right=0.5cm of w] (v) {$v$}; % v node
  
  % Connect the nodes
  \edge{u}{w}; 
  \edge{w}{v}; 
\end{tikzpicture}
\vspace{5mm}
\end{center}

It is then clear that $u$ and $v$ are not independent, since
\begin{align}
    p(u,v) = \int p(u) p(w|u) p(v|w) \mathrm{d} w
    = p(u)  \int p(w|u) p(v|w) \mathrm{d} w
    = p(u) q(v|u).
\end{align}
Perturbing $u$ will change the distribution/value of $w$ which will consequently affect that of $v$, meaning that $u$ causally affects $v$. This effect is mediated by $w$, and hence we call $w$ a mediator. 

\section{Dependence and Causation}

We can chain the rules that were defined between three variables; $u$, $v$ and $w$, in order to determine the dependence between two nodes, $u$ and $v$, within any arbitrary probabilistic graphical model given a set $Z$ of observed nodes. This procedure is called D-separation, and it tells us two things. First, we can check whether
\begin{align}
    u \indep_Z v.
\end{align}
More importantly, however, we get {\it all} open paths between $u$ and $v$. These open paths are conduits that carry {\it statistical} dependence between $u$ and $v$ regardless of whether there is a causal path between $u$ and $v$, where we define a causal path as an open {\it directed path} between $u$ and $v$.\footnote{
    Unlike a usual path, in a direct path, the directions of all edges must agree with each other, i.e., pointing to the same direction. 
}

Dependencies arising from open, non-causal paths are often casually referred to as `spurious correlation' or `spurious dependency'. When we are performing causal inference for the purpose of designing a causal intervention in the future, it is imperative to dissect out these spurious correlations and identify true causal relationship between $u$ and $v$. It is however unclear whether we want to remove all spurious dependencies or whether we should only remove spurious dependencies that are unstable, when it comes to prediction in machine learning. We will discuss more about this contention later in the course. 

In this course, we do not go deeper into D-separation. We instead stick to a simple setting where there are only three or four variables, so that we can readily identify all open paths and determine which are causal and which others are spurious.

\section{Causal Effects}

We have so far avoided defining more carefully what it means for a node $u$ to effect another node $v$ causally. Instead, we simply said $u$ effects $v$ causally if there is a directed path from $u$ to $v$ under the underlying data generating process. This is however unsatisfactory, as there are loopholes in this approach. The most obvious one is that some of those directed edges may correspond to a constant function. For instance, an extreme case is where the structural causal model is
\begin{align}
    &a \leftarrow f_a(u, \epsilon_a) \\
    &v \leftarrow f_v(a, \epsilon_v),
\end{align}
where $f_a(\cdot) = 0$. In this case, the edge from $u$ to $a$ is effectively non-existent, although we wrote it as if it existed. 
Rather, we want to define a causal effect of $u$ on $v$ by thinking of how perturbation on $u$ propagates over the data generating process and arrives at $v$. 

More specifically, we consider forcefully setting the variable $u$ (the cause variable) to an arbitrary value $\hat{u}$. This corresponds to replacing the following line in the structural causal model $G=(V,F,U)$
\begin{align}
    u \leftarrow f_u(\mathrm{pa}(u), \epsilon_u)
\end{align}
with 
\begin{align}
    u \leftarrow \hat{u}.
\end{align}
$\hat{u}$ can be a constant or can also be a random variable as long as it is not dependent on any other variables in the structural causal model. 

Once this replacement is done, we run this modified structural causal model $\overline{G}(\hat{u})=(\overline{V},\overline{F},U)$ in order to compute the following expected outcome:
\begin{align}
    \mathbb{E}_{U}[v_{\overline{\mathcal{G}}(\hat{u})}],
\end{align}
where $U$ is a set of exogenous factors (e.g. noise.) If $u$ does not affect $v$ causally, this expected outcome would not change (much) regardless of the choice of $\hat{u}$. 

As an example, assume $u$ can take either $0$ or $1$, as in treated or placebo. We then check the expected treatment effect on the outcome $v$ by
\begin{align}
    \mathbb{E}_{U}[v_{\overline{\mathcal{G}}(\hat{u}=1)}]
    -
    \mathbb{E}_{U}[v_{\overline{\mathcal{G}}(\hat{u}=0)}].
\end{align}
We would want this quantity to be positive and large to know that the treatment has a positive causal effect on the outcome. 

This procedure of forcefully setting a variable to a particular value is called a $\mathrm{do}$ operator. The impact of $\mathrm{do}$ is more starkly demonstrated if we consider a probabilistic graphical model rather than a structural causal model. Let $p_u(u | \mathrm{pa}(u))$ be the conditional distribution over $u$ in a probabilistic graphical model $G=(V,E,P)$. We  construct a so-called interventional distribution as
\begin{align}
    p(v | \mathrm{do}(u = \hat{u})),
\end{align}
which states that we are now forcefully setting $u$ to $\hat{u}$ instead of letting it be a sample drawn from the conditional distribution $p_u(u | \mathrm{pa}(u))$. That is, instead of 
\begin{align}
    u \sim u | \mathrm{pa}(u),
\end{align}
we do
\begin{align}
    u \leftarrow \hat{u}.
\end{align}

In other words, we replace the conditional probability $p_u(u | \mathrm{pa}(u))$ with
\begin{align}
    p_u(u | \mathrm{pa}(u)) = \delta(u - \hat{u}),
\end{align}
where $\delta$ is a Dirac measure, or replace all occurrences of $u$ in the conditional probabilities of $\mathrm{child}(u)$ with a constant $\hat{u}$, where 
\begin{align}
    \mathrm{child}(u) = 
    \left\{
    u' \in V |
    (u, v') \in E
    \right\}.
\end{align}
As a consequence, in the new modified graph $\overline{G}$, there is no edge coming into $u$ (or $\hat{u}$) anymore, i.e., $\mathrm{pa}(u) = \emptyset$. $u$ is now independent of all the other nodes {\it a priori}. 

Because $\mathrm{do}$ modifies the underlying data generating process, $p(v|u=\hat{u}; G)$ and $p(v|\mathrm{do}(u=\hat{u}); G) = p(v|u=\hat{u}; \overline{G})$ differ from each other. This difference signifies the separation between statistical and causal quantities. We consider this separation in some minimal cases.

\section{Case Studies}

\paragraph{An unobserved confounder and a direct connection.}

Consider the case where $w$ was not observed.

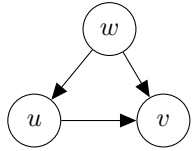
\begin{center}
\vspace{5mm}
\begin{tikzpicture}
  % Define nodes
  \node[latent] (u) {$u$}; % u node
  \node[latent, right=1cm of u] (v) {$v$}; % v node
  \node[latent, above=0.5cm of u, xshift=1cm] (w) {$w$}; % w node
  
  % Connect the nodes
  \edge{w}{u}; 
  \edge{w}{v};
  \edge{u}{v};
\end{tikzpicture}
\vspace{5mm}
\end{center}

Under this graph $G$,
\begin{align}
    p_G(v|u=\hat{u}) = \frac{\sum_{w} p(w) p(\hat{u}|w) p(v|\hat{u},w)}{p(\hat{u})} = \mathbb{E}_w \left[ \frac{p(\hat{u}|w)}{p(\hat{u})} p(v| \hat{u}, w) \right],
\end{align}
from which we see that there are two open paths between $v$ and $u$:
\begin{enumerate}
    \item $u \to v$: a direct path;
    \item $u \leftarrow w \to v$: an indirect path via the unobserved confounder.
\end{enumerate}

The statistical dependence between\footnote{
    We do not need to specify the direction of statistical dependence, since the Bayes' rule allows us to flip the direction.
} 
$u$ and $v$ flows through both of these paths, while the causal effect of $u$ and $v$ only flows through the direct path $u \to v$. The application of $\mathrm{do}(u = \hat{u})$ in this case would severe the edge from $w$ to $u$, as in

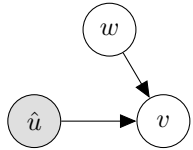
\begin{center}
\vspace{5mm}
\begin{tikzpicture}
  % Define nodes
  \node[obs] (u) {$\hat{u}$}; % u node
  \node[latent, right=1cm of u] (v) {$v$}; % v node
  \node[latent, above=0.5cm of u, xshift=1cm] (w) {$w$}; % w node
  
  % Connect the nodes
  % \edge{w}{u}; 
  \edge{w}{v};
  \edge{u}{v};
\end{tikzpicture}
\vspace{5mm}
\end{center}

Under this modified graph $\overline{G}$,
\begin{align}
    p_G(v | \mathrm{do}(u = \hat{u}))
    &=
    p_{\tilde{G}}(v | u=\hat{u})
    =
    \frac{1}{q(\hat{u})}\sum_w p(w) q(\hat{u}) p(v | \hat{u}, w)
    \\
    &=
    \sum_w p(w) p(v | \hat{u}, w)
    =
    \mathbb{E}_w \left[ p(v | \hat{u}, w) \right]
    ,
\end{align}
where we use $q(\hat{u})$ to signify that this is not the same as $p(\hat{u})$ above. 

The first one $p_G(v | u = \hat{u})$ is a statistical quantity, and we call it a conditional probability. The latter $p_G(v | \mathrm{do}(u = \hat{u}))$ is instead a causal quantity, and we call it an interventional probability. Comparing these two quantities, $p_G(v | u = \hat{u})$ and $p_G(v | \mathrm{do}(u = \hat{u}))$, the main difference is the multiplicative factor $\frac{p(\hat{u}|w)}{p(\hat{u})}$ inside the expectation. The numerator $p(\hat{u}|w)$ tells us how likely this treatment $\hat{u}$ was given under $w$, while the denominator $p(\hat{u})$ tells us how likely the treatment $\hat{u}$ was given overall. In other words, we upweight the impact of $\hat{u}$ and $w$ if $\hat{u}$ was more probable under $w$ than overall. This observation will allow us to convert between these two quantities later.

\paragraph{An observed collider and a direct connection.}

Let us flip the edges from $w$ so that those edges are directed toward $w$. We further assume that we always observe $w$ to be a constant ($1$).

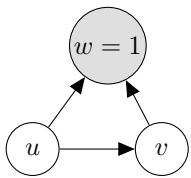
\begin{center}
\vspace{5mm}
\begin{tikzpicture}
  % Define nodes
  \node[latent] (u) {$u$}; % u node
  \node[latent, right=1cm of u] (v) {$v$}; % v node
  \node[obs, above=0.5cm of u, xshift=1cm] (w) {$w=1$}; % w node
  
  % Connect the nodes
  \edge{u}{w}; 
  \edge{v}{w};
  \edge{u}{v};
\end{tikzpicture}
\vspace{5mm}
\end{center}

The $\mathrm{do}$ operator on $u$ does not alter the graph above. This means that the conditional probability and interventional probability coincide with each other in this case, conditioned on observing $w=1$. 

This however does not mean that the conditional probability $p(v|u=\hat{u},w=1)$, 
or equivalently the interventional probability $p(v|\mathrm{do}(u=\hat{u}),w=1)$, measures the causal effect of $u$ on $v$ alone. As we saw before and also can see below, there are two open paths, $u \to v$ and $u \leftarrow w \to v$, between $u$ and $v$ through which the dependence between $u$ and $v$ flows:
\begin{align}
    p(v | u, w=1) = \frac{\cancel{p(u)} p(v | u) p(w=1|u, v)}{\sum_{v'} \cancel{p(u)} p(v'| u) p(w=1|u, v')}.
\end{align}

We must then wonder whether we can separate out these two paths from each other. It turned out unfortunately that this is not possible in this scenario, because we need the cases of $w \neq 1$ (e.g. $w=0$) for this separation. If you recall how we can draw samples from a probabilistic graphical model while conditioning some variables to take particular values, it was all about selecting a subset of samples drawn from the same graph without any observed variables. This selection effectively prevents us from figuring out the effect of $u$ on $v$ via $w$. We will have more discussion on this topic later in the semester in the context of invariant prediction.

Because of this inherent challenge, that may not even be addressable in many cases, we will largely stick to the case of having a confounder in this semester.

\section{Summary}

In this chapter, we have learned about the following topics:

\begin{enumerate}
    \item How to represent a data generating process: a probabilistic graphical model vs. a structural causal model;
    \item How to read out various distributions from a data generating process: ancestral sampling and Bayes' rule;
    \item The effect of confounders, colliders and mediators on independence;
    \item Causal dependency vs. spurious dependency;
    \item The $\mathrm{do}$ operator.
\end{enumerate}

\chapter{Active Causal Inference}

In this chapter, we assume the following graph $G$.  We use $a$, $y$ and $x$, instead of $u$, $v$ and $w$, to denote the action/treatment, the outcome and the covariate, respectively. The covariate $x$ is a confounder in this case, and it may or may not observed, depending on the situation.

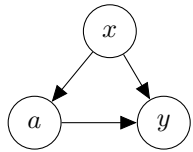
\begin{center}
\vspace{5mm}
\begin{tikzpicture}
  % Define nodes
  \node[latent] (a) {$a$}; % u node
  \node[latent, right=1cm of a] (y) {$y$}; % v node
  \node[latent, above=0.5cm of a, xshift=1cm] (x) {$x$}; % w node
  
  % Connect the nodes
  \edge{x}{a}; 
  \edge{x}{y};
  \edge{a}{y};
\end{tikzpicture}
\vspace{5mm}
\end{center}

An example case corresponding to this graph is vaccination. 

\begin{itemize}
    \item $a$: is the individual vaccinated?
    \item $y$: has the individual been infected by the target infectious disease with symptoms, within 6 months of vaccine administration?
    \item $x$: the underlying health condition of the individual.
\end{itemize}

The edge $x \to a$ is understandable, since we often cannot vaccinate an individual with an active underlying health condition. The edge $x \to y$ is also understandable, since healthy individuals may contract the disease without any symptoms, while immunocompromised individuals for instance may show a greater degree of symptoms with a higher chance. The edge $a \to y$ is also reasonable, as the vaccine must have been developed with the target infectious disease as its goal. In other words, this graphs encodes our structural prior about vaccination. 

With this graph that encodes the reasonable data generating process, causal inference then refers to figuring out the degree of the causal effect of $a$ on $y$.

\section{Causal Quantities of Interest}
\label{sec:ate}

In this particular case, we are interested in a number of causal quantities. The most basic and perhaps most important one is whether the treament is effective (i.e., results in a positive outcome) generally. This corresponds to checking whether $\mathbb{E}\left[y | \mathrm{do}(a=1)\right] > \mathbb{E}\left[y | \mathrm{do}(a=0)\right]$, or equivalently computing
\begin{align}
    \mathrm{ATE} = \mathbb{E}\left[y | \mathrm{do}(a=1)\right] - \mathbb{E}\left[y | \mathrm{do}(a=0)\right],
\end{align}
where 
\begin{align}
    \mathbb{E}\left[y | \mathrm{do}(a=\hat{a})\right] 
    &= 
    \sum_{y}
    y
    p(y | \mathrm{do}(a=\hat{a}))
    \\
    &=
    \sum_{y}
    y 
    \sum_{x} p(x) p(y|\hat{a}, x)
    =
    \sum_{y}
    y 
    \mathbb{E}_{x \sim p(x)} \left[
    p(y|\hat{a}, x)
    \right].
\end{align}

In words, we average the effect of $\hat{a}$ on $y$ over the covariate distribution but the choice of $\hat{a}$ should not depend on $x$. Then, we use this interventional distribution $p(y | \mathrm{do}(a=\hat{a}))$ to compute the average outcome. We then look at the difference in the average outcome between the treatment and not (placebo), to which we refer as the {\it average treatment effect} (ATE).

It is natural to extend ATE such that we do not marginalize the entire covariate $x$, but fix some part to a particular value. For instance, we might want to compute ATE but only among people in their twenties. Let us rewrite the covariate $x$ as a concatenation of $x$ and $x'$, where $x'$ is what we want to condition ATE on. That is, instead of $p(y | \mathrm{do}(a=\hat{a}))$, we are interested in $p(y | \mathrm{do}(a=\hat{a}), x'=\hat{x}')$. This corresponds to first modifying $G$ into
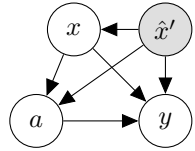
\begin{center}
\vspace{5mm}
\begin{tikzpicture}
  % Define nodes
  \node[latent] (a) {$a$}; % u node
  \node[latent, right=1cm of a] (y) {$y$}; % v node
  \node[latent, above=0.5cm of a, xshift=0.5cm] (x) {$x$}; % w node
  \node[obs, right=0.5cm of x] (xc) {$\hat{x}'$}; % w node
  
  % Connect the nodes
  \edge{x}{a}; 
  \edge{x}{y};
  \edge{xc}{x}; 
  \edge{xc}{a}; 
  \edge{xc}{y};
  \edge{a}{y};
\end{tikzpicture}
\vspace{5mm}
\end{center}

and then into
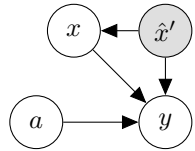
\begin{center}
\vspace{5mm}
\begin{tikzpicture}
  % Define nodes
  \node[latent] (a) {$a$}; % u node
  \node[latent, right=1cm of a] (y) {$y$}; % v node
  \node[latent, above=0.5cm of a, xshift=0.5cm] (x) {$x$}; % w node
  \node[obs, right=0.5cm of x] (xc) {$\hat{x}'$}; % w node
  
  % Connect the nodes
  \edge{x}{y};
  \edge{xc}{x}; 
  \edge{xc}{y};
  \edge{a}{y};
\end{tikzpicture}
\vspace{5mm}
\end{center}

We then get the following {\it conditional average treatment effect} (CATE):
\begin{align}
    \mathrm{CATE} = \mathbb{E}\left[y | \mathrm{do}(a=1), x'=\hat{x}'\right] - \mathbb{E}\left[y | \mathrm{do}(a=0), x'=\hat{x}'\right],
\end{align}
where
\begin{align}
    \mathbb{E}\left[y | \mathrm{do}(a=\hat{a}), x'=\hat{x}'\right] 
    &= 
    \sum_{y}
    y
    p(y | \mathrm{do}(a=\hat{a}), x'=\hat{x'})
    \\
    &=
    \sum_{y}
    y 
    \sum_{x} p(x|x') p(y|\hat{a}, x'=\hat{x'}, x)
    \\
    &=
    \sum_{y}
    y 
    \mathbb{E}_{x \sim p(x|x')} \left[
    p(y|\hat{a}, x'=\hat{x'}, x)
    \right].
\end{align}
You can see that this is really nothing but ATE conditioned on $x'=\hat{x}'$. 

From these two quantities of interest above, we see that the core question is whether and how to compute the interventional probability of the outcome $y$ given the intervention on the action $a$ conditioned on the context $x'$. Once we can compute this quantity, we can computer various target quantities under this distribution. We thus do not go deeper into other widely used causal quantities in this course but largely stick to ATE/CATE. 

\section{Regression: Causal Inference can be Trivial}
\label{sec:regression}

Assume for now that we are given a set of data points drawn from this graph $G$:
\begin{align}
    D = \left\{ (a_1, y_1, x_1), \ldots, (a_N, y_N, x_N) \right\}.
\end{align}
For every instance, we observe all of the action $a$, outcome $y$ and covariate $x$. Furthermore, we assume all these data points were drawn from the same fixed distribution 
\begin{align}
    p^*(a, y, x) = p^*(x) p^*(a|x) p^*(y|a,x)
\end{align} 
and that $N$ is large. 

In this case, we can use a non-parametric estimator, such as tables, deep neural networks and gradient boosted trees, to reverse-engineer each individual conditional distribution from this large dataset $D$. This is just like what we have discussed earlier in \S\ref{sec:learning}. Among three conditional distributions above, we are only interested in learning $p^*(x)$ and $p^*(y|a,x)$ from data, resulting $p(x;\theta)$ and $p(y|a,x;\theta)$, where $\theta$ refers to the parameters of each deep neural network.\footnote{
    Although there is no reason to prefer deep neural networks over random forests or other non-parametric learners, we will largely stick to deep neural networks, as I like them more. 
}

Once learning is over, we can use it to approximate ATE as
\begin{align}
    \mathrm{ATE} 
    &\approx 
    \sum_{y}
    y 
    \mathbb{E}_{x \sim p(x;\theta)} \left[
    p(y|a=1, x; \theta)
    \right]
    -
    \sum_{y}
    y 
    \mathbb{E}_{x \sim p(x;\theta)} \left[
    p(y|a=0, x; \theta)
    \right]
    \\
    &=
    \sum_{y}
    y 
    \mathbb{E}_{x \sim p(x;\theta)} \left[
    p(y|a=1, x; \theta) 
    -
    p(y|a=0, x; \theta)
    \right].
\end{align}

There are two conditions that make this regression-based approach to causal inference work:
\begin{enumerate}
    \item {\it No unobserved confounder}: we observe the covariate $x$ in $G$;
    \item Large $N$: we have enough examples to infer $p^*(y|a,x)$ with a low variance.
\end{enumerate}

If there is any dimension of $x$ that is not observed in the dataset, it is impossible for any learner to infer neither $p^*(y|a,x)$ nor $p^*(x)$ correctly. ``Correctly'' here in particular refers to identifying the true $p^*(y|a,x)$. This is not really important if the goal is to approximate the conditional probability $p^*(y|a)$, since we can simply drop $x$ and use $(a,y)$ pairs. It is however critical if the goal is to approximate the interventional probaiblity $p^*(y|\mathrm{do}(a))$ because this necessitates us to access $p^*(y|a,x)$ (approximately). 

Large $N$ is necessary for two reasons. First, the problem may be ill-posed when $N$ is small. Consider rewriting $p(y|a,x)$ as
\begin{align}
    p(y|a,x) = \frac{p(y,a,x)}{p(a|x)p(x)}.
\end{align}
This quantity has in the denominator both $p(a|x)$ and $p(x)$. If $N$ is small to the point that we do not observe all possible combination of $(a,x)$ for which $p(x) > 0$, this conditional probability is not well-defined. This connects to one of the major assumptions in causal inference, called {\it positivity}, which we will discuss further later in the semester.

The second, perhaps less important, reason is that the variance of the estimator is often inversely proportional to $N$. That is, with more $N$, we can approximate $p^*(y|a,x)$ with less variance. The variance of this estimate is critical, as it directly leads to that of ATE. If the variance of ATE is high, we cannot draw a confident conclusion whether the treatment is effective.

This section tells us that causal inference can be done trivially by statistical regression when the following conditions are satisfied:
\begin{enumerate}
    \item There are {\bf no unobserved confounder}: We observe every variable.
    \item {\bf Positivity}: All possible combinations of $(a,y,x)$ are observed in data.
    \item We have enough data.
\end{enumerate}

Unfortunately, it is rare that all these conditions are satisfied in real life. 

\section{Randomized Controlled Trials}
\label{sec:randomized-controlled-trials}

\subsection{The Basic Foundation}

In this section, we consider the case where there are unobserved confounders. In such a case, we cannot rely on regression, as these unobserved confounders prevent a learner from identifying $p^*(y|a, x)$ correctly. One may tempted to simply fit a predictive model on $(a, y)$ pairs to get $p(y|a; \theta)$ to approximate $p^*(y|a)$ and call it a day. We have however learned earlier that this does not approximate the causal effect, that is, $p^*(y|\mathrm{do}(a))$, due to the spurious path, $a \leftarrow x \to y$, which is open because we did not observe $x$. 

When there are unobserved confounders, we can actively collect data that allows us to perform causal inference. This is a departure from a usual practice in machine learning, where we often assume that data is provided to us and the goal is for us to use a learning algorithm to build a predictive model. This is often not enough in causal inference, and we are now presented with the first such case, where the assumption of `no unobserved confounder' has been violated. 

In order to estimate the causal effect of the action $a$ on the outcome $y$, we severed the edge from the confounder $x$ to the action $a$, as in

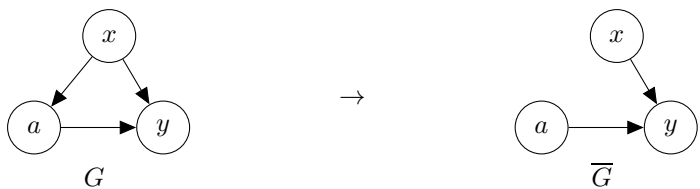
\begin{minipage}{0.4\textwidth}
\centering
\vspace{5mm}
\begin{tikzpicture}
  % Define nodes
  \node[latent] (a) {$a$}; % u node
  \node[latent, right=1cm of a] (y) {$y$}; % v node
  \node[latent, above=0.5cm of a, xshift=1cm] (x) {$x$}; % w node
  
  % Connect the nodes
  \edge{x}{a}; 
  \edge{x}{y};
  \edge{a}{y};
\end{tikzpicture}

$G$
\vspace{5mm}
\end{minipage}
\hfill
$\to$
\hfill
\begin{minipage}{0.4\textwidth}
\centering
\vspace{5mm}
\begin{tikzpicture}
  % Define nodes
  \node[latent] (a) {$a$}; % u node
  \node[latent, right=1cm of a] (y) {$y$}; % v node
  \node[latent, above=0.5cm of a, xshift=1cm] (x) {$x$}; % w node
  
  % Connect the nodes
  % \edge{x}{a}; 
  \edge{x}{y};
  \edge{a}{y};
\end{tikzpicture}

$\overline{G}$
\vspace{5mm}
\end{minipage}

This suggests that if we collect data according to the latter graph $\overline{G}$, 
we may be able to estimate the causal effect of $a$ on $y$ despite the unobserved confounder $x$. 
To do so, we need to decide on the prior distribution over the action, $\overline{p}(a)$, such that
$a$ is independent of the covariate $x$. It is a common practice to choose a uniform distribution over the action as $\overline{p}(a)$. For instance, if $a$ is binary,
\begin{align}
    \overline{p}(a=0) = \overline{p}(a=1) = 0.5.
\end{align}

The prior distribution over the covariate $x$, $p(x)$, is not what we choose, but is what the environment is like. 
That is, instead of specifying $p(x)$ nor sampling explicitly from it (which was by assumption impossible), we
go out there and recruit samples from this prior distribution. In the vaccination example above, this would correspond
to recruiting subjects from a general population without any particular filtering or selection.\footnote{
    Of course, we can filter these subjects to satisfy a certain set of criteria (inclusion criteria) in order to estimate a conditional average treatment effect.
} 

For each recruited subject $x$, we assign the treatment $a$, drawn from $\overline{p}(a)$ that is independent of $x$. 
This process is called `randomization', because this 
process assigns an individual drawn from the population, $x \sim p(x)$, randomly
to either a treatment or placebo group according to $\overline{p}(a)$, where `randomly' refer to `without any information'. 
This process is also `controlled', because we control the assignment of each individual to a group. 

For the randomly assigned pair $(a,x)$, we observe the outcome $y$ by letting the environment (nature) simulate and sample from $p^*(y|a,x)$. 
This process is a `trial', where we try the action $a$ on the subject $x$ by administering $a$ to $x$. Putting all these together, we call this process of collecting data from $\overline{G}$ a {\it randomized controlled trial} (RCT).

It is important to emphasize that $x$ is not recorded, fully known nor needed, and we end up with 
\begin{align}
    D = \left\{ (a_1, y_1), \ldots, (a_N, y_N) \right\}.
\end{align}
Using this dataset, we can now approximate the interventional probability as
\begin{align}
    \mathbb{E}_G\left[ y | \mathrm{do}(a=\hat{a})\right] 
    &= 
    \mathbb{E}_{\overline{G}} \left[ y | a = \hat{a}\right] 
    \\
    &=
    \sum_{y}
    y
    \frac{
    \sum_{x} p(x)
    \cancel{\overline{p}(\hat{a})}
    p(y | \hat{a}, x)}
    {\cancel{\overline{p}(\hat{a})}}
    \\
    &=
    \sum_{x}
    \sum_{y}
    y
    p(x)
    p(y | \hat{a}, x)
    \\
    &\approx
    \frac{\sum_{n=1}^N
    \mathds{1}(a_n = \hat{a})
    y_n}
    {\sum_{n'=1}^N \mathds{1}(a_{n'} = \hat{a})},
\end{align}
because $x_n \sim p(x)$, $a_n \sim \overline{p}(a)$ and $y_n \sim p(y|\hat{a}, x)$.

As evident from the final line, we do not need to know the confounder $x$, which means that 
RCT avoids the issue of unobserved, or even unknown, confounders. Furthermore, it does not
involve $\overline{p}(a)$, implying that we can use an arbitrary mechanism to randomly assign 
each subject to a treatment option, as long as we do not condition it on the confounder $x$. 
If we have strong confidence in the effectiveness of a newly developed vaccine, for instance, 
we would choose $\overline{p}(a)$ to be skewed toward the treatment ($a=1$).

\subsection{Important Considerations}

Perhaps the most important consideration that must be given when implementing a randomized controlled trial is to ensure that the action $a$ is independent of the covariate $x$. As soon as dependence forms between $a$ and $x$, the estimated causal effect $\mathbb{E}_G\left[ y | \mathrm{do}(a=\hat{a})\right]$ becomes biased. It is however very easy for such dependency to arise without a careful design of an RCT, often due to subconscious biases formed by people who implement the randomized assignment procedure. For instance, in the case of the vaccination trial above, a doctor may subconsciously assign older people less to the vaccination arm\footnote{
    An `arm' here refers to a `group'. 
}
simply because she is subconsciously worried that vaccination may have more severe side effects on an older population. Such a subconscious decision will create a dependency between the action $a$ and the covariate $x$ (the age of a subject in this case), which will lead to the bias in the eventual estimate of the causal effect. In order to avoid such a subconscious bias in assignment, it is a common practice to automate the assignment process so that the trial administrator is not aware of to which action group each subject was assigned. 

Second, we must ensure that the causal effect of the action $a$ on the outcome $y$ must stay constant throughout the trial. More precisely, $p^*(y|a,x)$ must not change throughout the trial. This may sound obvious, as it is difficult to imagine how for instance the effect of vaccination changes rapidly over a single trial. There are however many ways in which this does not hold true. A major way by which $p^*(y|a,x)$ drifts during a trial is when a participant changes their behavior. Continuing the example of vaccination above, let us assume that each participant knows that whether they were vaccinated and also that the pandemic is ongoing. Once the participant knows that they were given placebo instead of actual vaccine, they may become more careful about their hygiene, as they are worried about potential contracting the rampant infectious disease. Similarly, if they knew they were given actual vaccine, they may become less careful and expose themselves to more situations in which they could contract the disease. That is, the causal effect of vaccination changes due to the alteration of participants' behaviours. It is thus a common and important practice to blind participants from knowing to which treatment groups they were assigned. For instance, in the case of vaccination above, we would administer saline solution via injection to control (untreated) participants so that they cannot tell whether they are being injected actual vaccine or placebo.

Putting these two considerations together, we end up with a {\it double blind} trial design. In a double blind trial design, neither the participant nor the trial administrator is made aware of their action/treatment assignment. This helps ensure that the underlying causal effect is stationary throughout the study and that there is no bias creeping in due to the undesirable dependency of the action/treatment assignment on the covariate (information about the participant.) 

The final consideration is not about designing an RCT but about interpreting the conclusion from an RCT. As we saw above, the causal effect from the RCT based on $\overline{G}$ is mathematically expressed as
\begin{align}
    \mathbb{E}_G\left[ y | \mathrm{do}(a=\hat{a})\right] 
    =
    \sum_{x}
    \sum_{y}
    y
    p(x)
    p(y | \hat{a}, x).
\end{align}
The right hand side of this equation includes $p(x)$, meaning that the causal effect is conditioned on the prior distribution over the covariate. 

We do not have direct access to $p(x)$, but we have samples from this distribution, in the form of participants arriving and being included into the trial. That is, we have implicit access to $p(x)$. This implies that the estimated causal effect would only be valid when it is used for a population that closely follows $p(x)$. If there is any shift in the population distribution itself or there was any filtering applied to participants in the trial stage that does not apply after the trial, the estimated causal effect would not be valid anymore. For instance, clinical trials, such as the vaccination trial above, are often run by research-oriented and financially-stable clinics which are often located in affluent neighbourhoods. The effect of the treatment from such a trial is thus more precise for the population in such affluent neighbourhoods. This is the reason why inclusion is important in randomized controlled trials. 

Overall, a successful RCT requires the following conditions to be met:
\begin{enumerate}
    \item Randomization: the action distribution must be independent of the covariate.
    \item Stationarity of the causal effect: the causal effect must be stable throughout the trial.
    \item Stationarity of the popluation: the covariate distribution must not change during and after the trial.
\end{enumerate}
As long as these three conditions are met, RCT provides us with an opportunity to cope with unobserved confounders. 

\section{Causal Inference vs. Outcome Maximization}

Beside curiosity, the goal of causal inference is to use the inferred causal relationship for better outcomes in the future. Once we estimate $\mathbb{E}_G\left[ y | \mathrm{do}(a)\right]$ using RCT, we would simply choose the following action for all future subject:
\begin{align}
    \hat{a} = \arg\max_{a \in \mathcal{A}} \mathbb{E}_G\left[ y | \mathrm{do}(a)\right],
\end{align}
where $\mathcal{A}$ is a set of all possible actions. This approach has however one downside that we had to give an incorrect treatment (e.g. placebo) to many trial participants who lost their opportunities to have a better outcome (e.g. protection against the infectious disease.) 

Consider an RCT where subjects arrive and are tested serially, that is, one at a time. 
If $t$ subjects have participated in the RCT so far, we have
\begin{align}
    D = \left\{ (a_1, y_1), \ldots, (a_t, y_t) \right\}.
\end{align}
Based on $D$, we can estimate the outcome of each action by
\begin{align}
    \hat{y}_t(a) 
    =
    \frac{\sum_{t'=1}^t
    \mathds{1}(a_{t'} = a)
    y_{t'}}
    {\sum_{t''=1}^N \mathds{1}(a_{t''} = a)}.
\end{align}
This estimate would be unbiased (correct on average), if every $a_{t'}$ was drawn from an action distribution that is independent of the covariate $x$ from an identical distribution $q(a)$.\footnote{
    This is another constraint on RCT, that every subject must be assigned according to the same assignment policy $q(a)$. 
} 
More generally, the bias (the degree of incorrectness) would be proportional to 
\begin{align}
    \epsilon_{\leq t-1} = \frac{1}{t} 
    \sum_{t'=1}^t \mathds{1}(a_{t'} \text{ was drawn independently of } x_{t'}).
\end{align}
If $\epsilon_{\leq t-1} = 1$, the estimate is unbiased, corresponding to causal inference. 
If $\epsilon_{\leq t-1} = 0$, what we have is not interventional but conditional. 

Assuming $\epsilon_t \gg 0$ and $t \gg 1$, we have a reasonable causal estimate of the outcome $y$ given each action $a$. Then, in order to maximize the outcome of the next subject ($x_{t+1}$), we want to assign them to an action sampled from the following Boltzmann distribution:
\begin{align}
    q_t(a) = 
    \frac{\exp\left( \frac{1}{\beta_t} \hat{y}_t(a) \right)}
    {\sum_{a' \in \mathcal{A}} \exp\left( \frac{1}{\beta_t} \hat{y}_t(a') \right)},
\end{align}
where $\beta_t \in [0, \infty)$ is a temperature parameter.  

When $\beta_t \to \infty$ (a high temperature), this is equivalent to sampling the action $a_{t+1}$ from a uniform distribution, which
implies that we do not trust the causal estimates of the outcomes, perhaps due to small $t$. 
On the other hand, when $\beta_t \to 0$ (a low temperature), the best-outcome action would be selected, as
\begin{align}
    q_t(a) 
    =_{\beta \to \infty}
    \begin{cases}
        1, &\text{ if } \hat{y}(a) = \max_{a'} \hat{y}_t(a') \\
        0, &\text{ otherwise}
    \end{cases}
\end{align}
assuming there is a unique action that leads to the best outcome. In this case, we are fully trusting our causal estimates of the outcomes and simply choose the best action accordingly, which corresponds to {\it outcome maximization}.

We now combine these two in order to make a trade off between causal inference and outcome maximization. At time $t$, we sample the action $a_t$ for a new participant from
\begin{align}
\label{eq:bandit_policy}
    q_t(a)
    =
    \epsilon_t \frac{1}{|\mathcal{A}|}
    +
    (1-\epsilon_t)
    \frac{\exp\left( \frac{1}{\beta_t} \hat{y}_t(a) \right)}
    {\sum_{a' \in \mathcal{A}} \exp\left( \frac{1}{\beta_t} \hat{y}_t(a') \right)},
\end{align}
where $\epsilon_t \in [0, 1]$ and $|\mathcal{A}|$ is the number of all possible actions.

We can sample from this mixture distribution by
\begin{enumerate}
    \item Sample $e_t \in \left\{0,1\right\}$ from a Bernoulli distribution of mean $\epsilon_t$.
    \item Check $e_t$
    \begin{itemize}
        \item If $e_t=1$, we uniformly choose $a_t$ at random.
        \item If $e_t=0$, we sample $a_t$ proportionally to $\hat{y}(a)$.
    \end{itemize}
\end{enumerate}

As we continue the RCT according to this assignment policy, we assign participants increasingly more to actions with better outcomes, because our causal estimate gets better over time. We however ensure that participants are randomly assigned to actions at a reasonable rate of $\epsilon_t$, in order to estimate the causal quantity rather than the statistical quantity. It is common to start with a larger $\epsilon_t \approx 1$ and gradually anneal it toward $0$, as we want to ensure we quickly estimate the correct causal effect early on. It is also usual to start with a large $\beta_t \geq 1$ and anneal it toward $0$, as the early estimate of the causal effect is often not trustworthy. 

When the first component (the uniform distribution) is selected, we say that we are exploring, and otherwise, we say we are exploiting. $e_t$ is a hyperparameter that allows us to compromise between exploration and exploitation, while $\beta_t$ is how we express our belief in the current estimate of the causal effect. 

This whole procedure is a variant of EXP-3, which stands for the exponential-weight algorithm for exploration and exploitation~\citep{allesiardo2017non}, that is used to solve the {\it multi-armed bandit problem}. However, with an appropriate choice of $\epsilon_t$ and $\beta_t$, we obtain as a special case RCT that can estimate the causal effect of the action on the outcome. For instance, we can use the following schedules of these two hyperparameters:
\begin{align}
    \epsilon_t = \beta_t =
    \begin{cases}
        1,& \text{ if } t < T \\
        0,& \text{ if } t \geq T
    \end{cases}
\end{align}
with a larger $T \gg 1$. 
We can however choose smoother schedulers for $\epsilon_t$ and $\beta_t$ in order to make a better compromise between causal inference and outcome maximization, in order to avoid assigning too many subjects to a placebo (that is, ineffective) group. 

The choice of $\epsilon_t$ and $\beta_t$ also affects the bias-variance trade-off. Although this is out of the scope of this course, it is easy to guess that higher $\epsilon_t$ and higher $\beta_t$ lead to a higher variance but a lower bias, and vice versa.

\paragraph{A never-ending trial.}

A major assumption that must be satisfied for RCT is the stationarity. Both the causal distribution $p^*(y|a,x)$ and the covariate distribution $p^*(x)$ must be stationary in that they do not change throughout the trial as well as after the trial. Especially when these distributions drift after the trial, that is, after running the trial with $T$ participants, our causal estimate as well as the decision based on it will become less accurate. We see such instances often in the real world. For instance, as viruses mutate, the effectiveness of vaccination wanes over time, although the very same vaccine was found to be effective by an earlier RCT. 

When the underlying conditional distributions are all stationary, we do not need to keep the entire set of collected data points in order to compute the approximate causal effect, because
\begin{align}
    \hat{y}_t(a) 
    =
    \frac{\sum_{t'=1}^t
    \mathds{1}(a_{t'} = a)
    y_{t'}}
    {\sum_{t''=1}^N \mathds{1}(a_{t''} = a)}
    = 
    \frac{\sum_{t'=1}^{t-1}
    \mathds{1}(a_{t'} = a)}{\sum_{t'=1}^t
    \mathds{1}(a_{t'} = a)} \hat{y}_{t-1}(a)
    +
    \frac{
    \mathds{1}(a_{t} = a)}{\sum_{t'=1}^t
    \mathds{1}(a_{t'} = a)} 
    y_{t}.
\end{align}
In other words, we can just keep a single scalar $\hat{y}_t(a)$ for each action to maintain the causal effect over time. 

We can tweak this recursive formula to cope with slow-drifting underlying distributions by emphasizing recent data points much more so than older data points. This can be implemented with exponential moving average,\footnote{
    `exponential-weight' in EXP-3 comes from this choice. 
}
as follows:
\begin{align}
    \hat{y}_t(a) 
    =
    \begin{cases}
        \eta \hat{y}_{t-1}(a)
        + 
        (1-\eta) 
        y_t,& \text{ if }a_t = a
        \\
        \hat{y}_{t-1}(a),& \text{ if }a_t \neq a
    \end{cases}
\end{align}
where $\eta \in [0, 1)$. 
As $\eta \to 0$, we consider an increasingly smaller window into the past and do not trust what we have seen happen given a particular action. On the other hand, when $\eta \to 1$, we do not trust what happens now but rather what we already know about the causal effect of an action $a$ should be.

By keeping track of the causal effect with exponential moving average, we can continuously run the trial. When doing so, we have to be careful in choosing the schedules of $\epsilon_t$ and $\beta_t$. Unlike before, $\epsilon_t$ should not be monotonically annealed toward $0$, as earlier exploration may not be useful later when the underlying distributions drift. $\beta_t$ also should not be annealed toward $0$, as the estimate of the causal effect we have at any moment cannot be fully trusted due to the unanticipated drift of underlying distributions. It is thus reasonable to simply set both $\beta_t$ and $\epsilon_t$ to reasonably large constants.

\paragraph{Checking the bias.}

At time $t$, we have accumulated
\begin{align}
    D_t = \left\{ (e_1, a_1, y_1), \ldots, (e_t, a_t, y_t) \right\},
\end{align}
where $e_{t'}$ indicates whether we explored ($1$) or exploited ($0$) at time $t'$.

We can then get the unbiased estimate of the causal effect of $a$ on $y$ by only using the triplets $(e,a,y)$ for which $e=1$. That is,
\begin{align}
    \tilde{y}_t(a) 
    =
    \frac{\sum_{t'=1}^t
    \mathds{1}(e_{t'} = 1)
    \mathds{1}(a_{t'} = a)
    y_{t'}}
    {\sum_{t''=1}^t 
        \mathds{1}(e_{t''} = 1)
        \mathds{1}(a_{t''} = a)
    }.
\end{align}
This estimate is unbiased, unlike $\hat{y}(a)$ from EXP-3 above, since we only used the action-outcome pairs when the action was selected randomly from the same uniform distribution. 

Assuming a large $t$ (so as to minimize the impact of a high variance,) we can then compute the (noisy) bias of the causal effect estimated and used by EXP-3 above as
\begin{align}
    b_t = (\hat{y}_t(a) - \tilde{y}_t(a))^2.
\end{align}
Of course this estimate of the bias is noisy and especially so when $t$ is small, since the effective number of data points used to estimate $\tilde{y}_t$ is on average
\begin{align}
    \sum_{t'=1}^t \epsilon_{t'} \leq t.
\end{align}

\section{When Some Confounders are Observed}

The assignment of an individual $x$ to a particular treatment option $a$ is called a policy. In the case of RCT, this policy was a uniform distribution over all possible actions $a$, and in the case of EXP-3, it was a mixture of a uniform policy and a effect-proportional policy from Eq.~\eqref{eq:bandit_policy}. In both cases, the policy was not conditioned on the covariate $x$, meaning that no information about the individual was used for assignment. This is how we addressed the issue of unobserved confounders. 

Such an approach is however overly restrictive in many cases, as some treatments may only be effective for a subset of the population that share a certain trait. For instance, consider the problem of inferring the effect of a monoclonal antibody therapeutics called Trastuzumab (or Herceptin) for breast cancer on the disease-free survival of a patient~\citep{nahta2007trastuzumab}. If we run RCT without taking into account any covariate information as above, we very likely would not see any positive effect on the patient's disease-free survival, because Trastuzumab was specifically designed to work for HER2-positive breast cancer. That is, only breast cancer patients with over-expressed ERBB2 gene (which encodes the HER2 receptor) would benefit from Trastuzumab. In these cases, we are interested in conditional average treatment effect (CATE) from earlier, that is, to answer the question of what causal effect Trastuzumab has on patients given their gene expression profile. CATE given the overly-expressed ERBB2 gene of Trastuzumab would be positive, while CATE without over-expression of ERBB2 gene would be essentially zero.

When we observe some confounders, such as the gene expression profile of a subject in the example above, and do not observe all the other confounders, we can mix RCT (\S\ref{sec:randomized-controlled-trials}) and regression (\S\ref{sec:regression}) to estimate the conditional causal effect conditioned on the observed confounders. 

The graph $G$ with the partially-observed confounder is depicted as below with the unobserved $x$ and the observed $x'$:
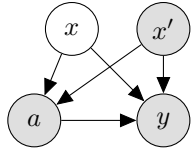
\begin{center}
\vspace{5mm}
\begin{tikzpicture}
  % Define nodes
  \node[obs] (a) {$a$}; % u node
  \node[obs, right=1cm of a] (y) {$y$}; % v node
  \node[latent, above=0.5cm of a, xshift=0.5cm] (x) {$x$}; % w node
  \node[obs, right=0.5cm of x] (xc) {$x'$}; % w node
  
  % Connect the nodes
  \edge{x}{a}; 
  \edge{x}{y};
  \edge{xc}{a}; 
  \edge{xc}{y};
  \edge{a}{y};
\end{tikzpicture}
\vspace{5mm}
\end{center}

Each subject in RCT then corresponds to the action-outcome-observed-covariate triplet $(a_t, y_t, x'_t)$. Assume we have enrolled and experimented with $t$ participants so far, resulting in 
\begin{align}
    D_t = \left\{ (a_1, y_1, x'_1), \ldots, (a_t, y_t, x'_t) \right\}.
\end{align}
Let $\hat{x}'$ be the condition of interest, such as the overexpression of ERBB2. we can then create a subset $D(\hat{x}')$ as
\begin{align}
    D_t(\hat{x}') = \left\{
    (a, y, x') \in D_t
    |
    x' = \hat{x}'
    \right\} \subseteq D_t.
\end{align}

We can then use this subset $D_t(\hat{x}')$ as if it were the set from the ordinary RCT, in order to estimate the conditional causal effect, as follows.
\begin{align}
    \hat{y}_t(a|x')
    \approx
    \frac{
    \sum_{(a_i,y_i,x_i') \in D_t(x')}
    \mathds{1}(a_i=a)
    y_i
    }
    {
    \sum_{(a_j,y_j,x_j') \in D_t(x')}
    \mathds{1}(a_j=a)
    }.
\end{align}
Just like what we did earlier, we can easily turn this into a recursive version in order to save memory:
\begin{align}
\hat{y}_t(a|x')
=
\begin{cases}
    \hat{y}_{t-1}(a|x'),&\text{ if } x'_t \neq x' \\
    \frac{
    \hat{y}_{t-1}(a|x')
    \sum_{(a_j,y_j,x_j') \in D_{t-1}(x')}
    \mathds{1}(a_j=a)
    +
    y_t
    \mathds{1}(a_t = a)
    }
    {
    \sum_{(a_k,y_k,x_k') \in D_{t}(x')}
    \mathds{1}(a_k=a)
    },&\text{ if }x'_t = x'
\end{cases}
\end{align}
Similarly to earlier, we can instead of exponential moving average in order to cope with the distribution drift over the sequential RCT, as 
\begin{align}
\hat{y}_t(a|x')
=
\begin{cases}
    \hat{y}_{t-1}(a|x'),&\text{ if } x'_t \neq x' \\
    \eta_t 
    \hat{y}_{t-1}(a|x')
    +
    (1-\eta_t)
    y_t
    \mathds{1}(a_t = a),
    &\text{ if }x'_t = x'
\end{cases}
\end{align}
We can then use this running estimates of the causal effects in order to build an assignment policy $\pi_t(a|x')$ that now depends on the observed covariate $x'$. If we go back to the earlier example of Trastuzumab, this policy would increasingly more assign participants with over-expressed ERBB2 to the treatment arm, while it would continue to be largely uniform for the remaining population. 

\paragraph{A Parametrized Causal Effect.}

Up until this point, partially-observed confounders do not look like anything special. It is effectively running multiple RCT's in parallel by running an individual RCT for each covariate configuration. There is no benefit of running these RCT's in parallel relative to running these RCT's in sequence. We could however imagine a scenario where the former is more beneficial than the latter, and we consider one such case here.

Assume that $x'$ is a multi-dimensional vector, i.e., $x' \in \mathbb{R}^d$ and that the true causal effect of each action $a$ is a linear function of the observed covariate $x'$:
\begin{align}
    \hat{y}^*(a|x') = \theta^*(a)^\top x' + b^*(a) = \sum_{i=1}^d \theta_d^*(a) x'_d + b(a).
\end{align}
This means that each dimension $x'_d$ of the covariate has an additive effect on the expected outcome $\hat{y}(a|x)$ weighted by the associated coefficient $\theta_d^*(a)$, and that the effect $\theta_d^*(a) x'_d$ of each dimension on the expected outcome is independent of the other dimensions' effects. 

As an example, consider estimating the effect of weight lifting on the overall health. 
The action is whether to perform weight lifting each day, and the outcome is the degree of the subject's healthiness. Each dimension $x'_d$ refers to a habit of a person. For instance, it could be a habit of smoking, and the corresponding dimension $x'_d$ encodes the number of cigarettes the subject smokes a day. Another habit could be jogging, and the corresponding dimension would encode the number of minutes the subject runs a day. 
Smoking is associated with a negative coefficient regardless of $a$. On the other hand, jogging is associated with a negative coefficient when $a=1$, because an excessive level of workout leads to frequent injuries, while it is with a positive coefficient when $a=0$. 

Of course, some of these habits may have nonlinear effects. Running just the right duration each day in addition to weight lifting could lead to a better health outcome. It is however reasonable to assume linearity as the first-order approximation. 

We can estimate the coefficients $\theta(a)$ by regression from \S\ref{sec:regression} by solving
\begin{align}
\min_{\theta(a)}
    \frac{1}{2} 
    \sum_{t'=1}^t
    \mathds{1}(a_{t'} = a)
    \left( 
    y_{t'} - \theta(a)^\top x'_{t'} - b^*(a)
    \right)^2.
\end{align}
Instead of keeping the count for each and every possible $x'$, we now keep only $\theta(a)$ for each action $a$. This has an obvious advantage of requiring only $O(d)$ memory rather than $O(2^d)$. 

More importantly however is that the estimated causal effect generalizes unseen covariate configuration. Let us continue from the example of having smoking and jogging as two dimensions of $x'$. During RCT, we may have seen participants who either smoke or jog but never both. Because of the linearity, the estimated causal effect predictor,
\begin{align}
    \theta(a)_{\mathrm{smoke}} x'_{\mathrm{smoke}} + \theta(a)_{\mathrm{run}} x'_{\mathrm{run}},
\end{align}
generalizes to participants who both smokes and jogs as well as who neither smokes nor jogs. 

This case of a linear causal effect suggests that we can rely on the power of generalization in machine learning in order to bypass the strong assumption of positivity. Even if we do not observe a covariate or an associated action, a parametrized causal effect predictor can generalize to those unseen cases. We will discuss this potential further later in the semester.

\paragraph{When there are many possible actions.}

Assume we do not observe any confounder, that is, there is no $x'$. Then, at each time, RCT is nothing but estimating a single scalar for each action. Let $\mathcal{A}$ be a set of all actions and $|\mathcal{A}|$ a cardinality of this action set. Then, at any time $t$ of running an RCT, the number of data points we can use to estimate the causal effect of a particular action is
\begin{align*}
    N(a) = \sum_{t'=1}^t \mathds{1}(a_{t'} = a) \approx t p_a(a),
\end{align*}
where $p_a(a)$ is the probability of selecting the action $a$ during randomization. Just like the case above with partially observed confounders, the variance of the estimate of the causal effect of an individual action decreases dramatically as the number of possible actions increases. 

We must have some extra information (context) about these actions in order to break out of this issue. Let $c(a) \in \mathbb{R}^d$ be the context of the action $a$. For instance, each dimension of $c(a)$ corresponds to the amount of one ingredient for making the perfect steak seasoning, such as salt, pepper, garlic and others. Then, each action $a$ corresponds to a unique combination of these ingredients. 

In this case, the causal effect of any particular action can be thought of mapping $c(a)$ to the outcome $\hat{y}(a)$ associated with $a$. If we assume this mapping was linear~\citep{li2010contextual}, we can write it as
\begin{align}
    \hat{y}(a) =
    c(a)^\top {\theta^*}  + b^*, 
\end{align}
where $\theta \in \mathbb{R}^d$ and $b \in \mathbb{R}$. 

Similarly to the case where there was an observed confounder above, with linearity, we do not need to maintain the causal estimate for each and every possible action, which amounts to $|\mathcal{A}|$ numbers, but the effect of each dimension of the action context on the outcome, which amounts of $d$ numbers. When $d \ll |\mathcal{A}|$, we gain a significant improvement in the variance of our estimates.

Furthermore, just like what we saw above, we benefit from the compositionality, or compositional generalization. For instance, if the effects of salt and pepper on the final quality of seasoning are independent and additive, we can accurately estimate the effect of having both salt and pepper even when all tested seasonings had either salt or pepper but never both. Let $c(a)=[s_{\mathrm{salt}}, s_{\mathrm{pepper}}]$, and assume $s_{\mathrm{salt}}, s_{\mathrm{pepper}} \in \{0, 1\}$ and that all past trials were such that $s_{\mathrm{salt}}=0$ or $s_{\mathrm{pepper}} = 0$. We can approximate $\theta^*_{\mathrm{salt}}$ and $\theta^*_{\mathrm{pepper}}$ from these past trials, and due to the linearity assumption, we can now compute the causal effect of $c(a)=[1, 1]$, as
\begin{align}
    \hat{\theta}_{\mathrm{salt}} + \hat{\theta}_{\mathrm{pepper}}.
\end{align}
This would not have been possible without the linearity, or more generally compositionality, because this particular action of adding both salt and pepper has never been seen before, i.e., it violates the positivity assumption. This is yet another example of overcoming the violation of positivity by generalization. 

At this point, one sees a clear connection between having some confounders observed and having many actions associated with their contexts. This is because they are simply two sides of the same coin. I leave this to you to think of why this is the case.

\section{Summary}

In this chapter, we have learned about the following topics:

\begin{enumerate}
    \item Average treatment effect; 
    \item Regression for causal inference;
    \item Randomized controlled trials;
    \item Outcome maximization with a bandit algorithm;
    \item A contextual bandit.
\end{enumerate}

\chapter{Passive Causal Inference}

\section{Challenges in Randomized Controlled Trials}

A major issue with randomized controlled trials (RCT) is that we must experiment with subjects. This raises many issues that are not necessarily related to causal inference itself but are more broadly about ethics and legality. For instance, the ``Tuskegee Study of Untreated Syphilis in the Negro Male'' was the widely-known and widely-condemned study for investigating the effect of untreated syphilis~\citep{centers2020tuskegee}. As RCT requires careful, double blinding, the trial administrators did not reveal to the study participants that they were diagnosed with (latent) syphilis. The study was originally designed (and the participants were told) to run for six months but lasted for 40 years until the details of the study were leaked to the press. During these decades, the treatment for syphilis was made available but none of the participants were treated properly, resulting in the death of more than 100 participants due to syphilis, out of approximately 400 participants, the syphilis infection of the wives of fourty participants and the congenital syphilis infection of 19 children. It took more than half a century for the US government to formally issue apology. 

A similar issue persists throughout medicine when it comes to RCT which is {\it de facto} standard for establishing any causal effect of a treatment on the outcome of a patient. Due to the necessity of randomization, some patient participants will inevitably receive placebo rather than the actual treatment. Even if the tested treatment ultimately turns out to be causally effective, by then it may be already late for those patients who were put on the control arm to receive and benefit from this new treatment. How ready are you to put patients into suffering because we want to (and often need to) establish the causal effect of a new treatment? 

Sometimes, it is impossible to design a placebo that ensures double blindness of a trial. Consider for instance an RCT on the effectiveness of masking on preventing respiratory diseases. Participants will understandably alter their behaviours based on their assignments; treatment (masking) or control (no masking), as their perception of risk is altered, which violates the stationarity of the causal effect $p^*(y|a,x)$. In order to avoid this, we must ensure that participants cannot tell whether they are in the treatment or control arm, but it is pretty much impossible to design a placebo mask that looks and feels the same as an actual mask but does not filter any particle in the air. In other words, RCT is only possible when placebos can be effectively designed and deployed. 

In this example, we run into yet another problem; how do we enforce the treatment on subjects? In the case of vaccination, subjects come into clinics and are for instance injected on the spot under the supervision of a clinician, after which the subjects cannot get rid of injected vaccine. In the case of masking, for instance, we cannot ensure that participants wear masks as they are instructed, as this requires non-stop monitoring throughout the trial period. 

Finally, some actions take long to have measurable impact on the outcome. For instance, consider a policy proposal of introducing a new course on programming at elementary schools (1-6 grades) with the goal of improving students' job prospects and growing the information technology (IT) sector. It will take anywhere between 12 to 20 years for these students to finish their education and participate in society, and we will have to wait another 4 to 15 years to see any measurable economic impact on the IT sector. Such a long duration between the action and the outcome further complicates RCT, as it is often impossible to ensure the stationarity of underlying distributions over that duration. RCT is thus not suitable for such actions that require a significant amount of time to have any measurable impact.

In this chapter, we instead consider an alternative approach to RCT, where we rely on existing data to infer the causal relationship between the action and outcome. As we use already collected data, we can often avoid the issues arising from actively experimentation, although we are now faced with another set of challenges, such as the existence of spurious correlations arising from various unobserved confounders that affected the choice of actions earlier. We will discuss how we can avoid these issues in this chapter. It is however important to emphasize that there is no silver bullet in causal inference. 

\section{When Confounders were also Collected}

\subsection{Inverse Probability Weighting}
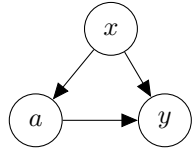
\label{sec:ipw}

\begin{center}
\begin{tikzpicture}
  % Define nodes
  \node[latent] (a) {$a$}; % u node
  \node[latent, right=1cm of a] (y) {$y$}; % v node
  \node[latent, above=0.5cm of a, xshift=1cm] (x) {$x$}; % w node
  
  % Connect the nodes
  \edge{x}{a}; 
  \edge{x}{y};
  \edge{a}{y};
\end{tikzpicture}
\end{center}

Let us come back to the original graph $G$ that consists of three variables; $a$, $y$ and $x$, with the following joint probability:
\begin{align}
    p^*(a, y, x) = p^*(x) p^*(a|x) p^*(y|a, x).
\end{align}
We also assume that we have a set $D$ of triplets $(a_n, y_n, x_n)$ drawn from this underlying graph $G$. In other words, we assume that we observe the confounder in this case. 

If we have a large such set, i.e. $N=|D| \gg 1$, we can use regression, as in \S\ref{sec:regression}, to approximate $p^*(y | a, x)$ in order to compute the causal effect as
\begin{align}
\label{eq:regression}
    \mathbb{E}_G\left[ y | \mathrm{do}(a=\hat{a})\right] 
    &\approx
    \sum_{x}
    p^*(x)
    \sum_{y}
    y
    \hat{p}(y | \hat{a}, x)
    \\
    &\approx
    \frac{1}
    {N}
    \sum_{n=1}^N
    \mathbb{E}_{\hat{p}(y | \hat{a}, x_n)}[y].
\end{align}

Although this regression-based approach is straightforward, this approach has a disadvantage of having to fit a regressor on the concatenation of the action $a$ and confounder $x$. The issue is exactly that of RCT with partially observed confounders, that is, we must have enough data points for each action-covariate combination in order for regression to have a low variance. We can use regularization to reduce the variance, which may unfortunately introduce a bias. 

We can reduce the variance by using regression to estimate a simpler quantity. In particular, we consider approximating $p^*(a|x)$. Because this is a map from $\mathcal{X}$, we just need enough data points for each covariate configuration rather than the action-covariate combination. Approximating $p^*(a|x)$ allows us to estimate the causal effect using data points drawn from the original graph $G$ rather than the modified graph $\overline{G}$ from \S\ref{sec:randomized-controlled-trials}, because
\begin{align}
    \mathbb{E}_G\left[ y | \mathrm{do}(a=\hat{a})\right] 
    &=
    \sum_{x}
    p^*(x)
    \sum_{a}
    \mathds{1}(a = \hat{a})
    \sum_{y}
    p^*(y | \hat{a}, x)
    y
    \\
    &= 
    \sum_{x}
    \sum_{a}
    \sum_{y}
    p^*(x)
    p^*(a|x)
    \mathds{1}(a = \hat{a})
    p^*(y | \hat{a}, x)
    \frac{1}{p^*(a|x)}
    y
    \\
    &=
    \frac{1}{\sum_{n'=1}^N
    \mathds{1}(a_{n'}= \hat{a})}
    \sum_{n=1}^N
    \mathds{1}(a_n = \hat{a})
    \frac{y_n}
    {p^*(\hat{a} | x_n)}.
\end{align}

Instead of the true $p^*(\hat{a}|x_n)$, we plug in the regression-based approximation $\hat{p}(\hat{a} | x_n)$ and arrive at 
\begin{align}
    \label{eq:ipw}
    \mathbb{E}_G\left[ y | \mathrm{do}(a=\hat{a})\right] 
    &
    \approx
    \frac{\sum_{n=1}^N
    \mathds{1}(a_n = \hat{a})
    \frac{y_n}
    {\hat{p}(\hat{a} | x_n)}}{
    \sum_{n'=1}^N
    \mathds{1}(a_{n'}= \hat{a})}.
\end{align}

In words, we look for all data points within the previously-collected set $D$ that are associated with the action $\hat{a}$ of interest. The simple average of the associated outcomes would be a biased estimate of the casual effect, since it combines the effects of $a$ on $y$ via two paths; (causal) $a \to y$ and (spurious) $a \leftarrow x \to y$. We correct this bias by weighting each data point, or the associated outcome, by the inverse probability of the action given the confounder, $\frac{1}{\hat{p}(\hat{a} | x_n)}$. This approach is thus called {\it inverse probability weighting} (IPW), and $p^*(a|x)$ is often referred to as a propensity score. 

\paragraph{It is fine to have missing outcomes.}

One important advantage of the IPW-based approach is that we can approximate $p^*(a|x)$ using all $(a,x)$ pairs even if they are not associated with the outcome $y$, unlike the earlier regression-based approach which required having all three variables observed $(a,y,x)$. Imagine using clinical notes and measurements from the electronic health record (EHR) of a large hospital in order to estimate the causal effect of a particular drug on a target disease. Of course, the prescription of the drug $a$ is not made blindly but based on the patient information which includes their underlying health condition. Since the existing health conditions affect the outcome $y$ of almost any kind of a disease, such patient information is a confounder $x$. Some patients often do not return to the same hospital for follow-up checks, meaning that the EHR does not record the outcome of these patients, leaving us only with $(a,x)$. We can then use all the prescriptions to approximate the propensity score $p(a|x)$ and then use a subset of these prescriptions for which the patients' outcomes are recorded (i.e. they came back for follow-up visits) to compute the causal effect. 

Mathematically, it means that we solve two separate optimization problems using two different data sets (though, one is a superset of the other,) as follows: 
\begin{align}
    &\hat{p}(a|x) = \arg\max_{p} \sum_{(a',x') \in D_{\pi}} \log p(a'|x'),
    \\
    &\hat{y}(\hat{a}) =   
    \frac{\sum_{(a',x',y') \in D_{\bar{y}}}
    \mathds{1}(a' = \hat{a})
    \frac{y_n}
    {\hat{p}(\hat{a} | x_n)}}{
    \sum_{(a'',x'',y'') \in D_{\bar{y}}}
    \mathds{1}(a'' = \hat{a})
    },
    \label{eq:ipw}
\end{align}
where $D_{\bar{y}} \subseteq D_{\pi}$. 

% \paragraph{Known policies.}

% We sometimes have access to the original policy $p^*(a|x)$ by which the action was assigned to each individual. Major medical associations release various policies by which diagnosis as well as treatment planning should happen for various diseases, and these policies can be used directly in place of the propensity score. This completely avoids the necessity of data collected for inferring the original policy.

\paragraph{A doubly robust estimator.}

A major issue with the IPW-based approach is that the variance of the estimator can be very large, even if the variance of estimating the propensity score is low, because the propensity scores shows up in the denominator. On the other hand, the regression-based approach has a low variance due to the missing division by the propensity score. It is however likely a biased estimator, as we cannot easily guarantee that the choice of a non-parametric regressor can identify the correct conditional probability, due to a variety of issues, such as the lack of realizability. 

If the regression-based approach is correct for an instance $(a,y,x)$, $\tilde{y}(a, x)$ would coincide with $y$, and we would just use $\tilde{y}(a, x)$ as it is and prefer to avoid the IPW-based estimate due to the potentially large variance arising from the denominator. Otherwise, we want to rely on the IPW-based estimate to correct for the incorrectness arising from the regression-based approach. This can be expressed as
\begin{align}
    \hat{y}(\hat{a})
    =
    \sum_x p(x)
    \sum_a
    \left(
    \frac{1}{|\mathcal{A}|} 
    \tilde{y}(a,x)
    +
    \underbrace{
    p^*(a|x)
    \frac{\tfrac{1}{|\mathcal{A}|}}{\hat{p}(a|x)}
    }_{(b)}
    \left(
    \underbrace{
    y^*(a|x)
    -
    \tilde{y}(a,x)
    }_{\mathrm{(a)}}
    \right)
    \right).
\end{align}
If $\tilde{y}(a,x)$ is perfect, (a) disappears, as expected. If our estimate of the propensity score is perfect, (b) is $1$, resulting in using the true $y^*(a,x)$ while ignoring the regression-based approach.\footnote{
    $y^*(a,x)$ is the true expected outcome given the action $a$ and the covariate $x$. Since expectation is linear, we can push the expectation all the way inside to obtain this expression. 
} 

Since we are provided with data rather than the actual probability distributions, we end up with 
\begin{align}
    \hat{y}(\hat{a}) 
    =  
    \frac{1}{Z(\hat{a})}
    \sum_{(a',x',y') \in D}
    \mathds{1}(a' = \hat{a})
    \left(
    \tilde{y}(\hat{a}, x_n)
    +
    \frac{1}
    {\hat{p}(\hat{a} | x_n)}
    \left(y_n - \tilde{y}(\hat{a}, x_n)\right)
    \right),
\end{align}
where $Z(\hat{a}) = \sum_{(a'',x'',y'') \in D}
    \mathds{1}(a'' = \hat{a}).$ This estimator is called a {\it doubly robust estimator}.

\subsection{Matching.}
\label{sec:matching}

Instead of estimating $p^*(a|x)$ and multiplying the observed outcome with its inverse, we can achieve a similar outcome by manipulating data itself. When $p^*(a|x) = p^*(a)$, that is, the action is independent of the covariate, the IPW-based estimate coincides with simple averaging, just like in RCT from \S\ref{sec:randomized-controlled-trials}. This happens when the ratio of actions associated with each unique $x$ in the data is the same across the data set. To make it simpler, let us express this ratio of actions by assigning the minimal number of each action in relation to the other actions. For instance, if our desired ratio between the treatment and placebo is $0.8:0.2$, we would express it for instance as $4:1$. 

Starting from the original data set $D=\left\{(a_1, x_1, y_1), \ldots, (a_N, x_N, y_N) \right\}$, we go through each $x_n$ by collecting as many $(a,x_n,y) \in D$ as $n_a$, where $n_a$ is the target number of examples of action $a$, for instance $4$ above. This collection can be done randomly or by following some fixed strategy, such as round-robin scheduling. Sometimes it is necessary to choose the same triplet multiple times, which is not ideal but may be necessary. By aggregating these collections, we arrive at a new dataset $\tilde{D}$. 

Under this new dataset $\tilde{D}$, the propensity score is guaranteed to be
\begin{align}
    \hat{p}(a|x) \propto n_a,
\end{align}
regardless of $x$. Furthermore, $\hat{p}(a|x) = \hat{p}(a)$ as well. Meanwhile, $\hat{p}(x)$ stays the same as that under the original data set $D$. Then, the expected causal outcome of any given action $a$ under this dataset is
\begin{align}
    \hat{y}(\hat{a}) = \frac{1}{N} \sum_{n=1}^N \mathds{1}(a'_n=\hat{a}) y'_n,
\end{align}
where $(a'_n, y'_n, x'_n) \in \tilde{D}$. This avoids the issue of the high variance arising from the IPW-based approach. This however does not mean that this approach always works. 

The most obvious issue with this approach is that the original data set may not have enough triplets associated with each $x$ to ensure that $p^*(a|x)$ is identical for all $x$. Furthermore, even if we have enough associated triplets for each $x$, we may end up with discarding many triples from the original data set to form the new data set. We never want to discard data points we have. This approach is thus appropriate when data is already well balanced and the goal is to further ensure that the propensity score is constant. 

A relatively simple variant of this approach, called `matching' because we match triplets based on $x$, is to relax the constraint that we only consider the exact match of the covariate $x$. The original formulation samples the triplets according to the target counts with replacement from the following multiset:\footnote{
    It is a multiset, since there can be duplicates. 
}
\begin{align}
    D(x) = \left\{ 
    (a',y',x') \in D |
    x' = x
    \right\}.
\end{align}
This condition of $x' = x$ may be too strict, leaving us with only a very small $D(x)$. 
We can instead relax this condition using a predefined notion of similarity such that
\begin{align}
    \tilde{D}(x) = \left\{ 
    (a',y',x') \in D |
    s(x',x) < \epsilon
    \right\},
\end{align}
where $s(x',x)$ is a predefined distance function and $\epsilon$ a distance threshold.

\section{Instrumental Variables: When Confounders were not Collected}
\label{sec:instrumental-variables}

So far in this section we have considered a case where the confounder $x$ was available in the observational data. This allowed us to either fit the regressor directly on $p(y|a,x)$, use inverse probability weighting or re-balance the dataset using the matching scheme. It is however unlikely that we are given full access to the confounders (or any kind of covariate) in the real world. It is thus important to come up with an approach that works on passively collected data without covariates.

\paragraph{An instrumental variable estimator.}

Let us rewrite the following graph $G_0$ into a corresponding structural causal model:

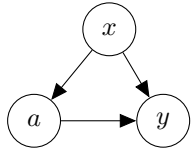
\begin{center}
\begin{tikzpicture}
  % Define nodes
  \node[latent] (a) {$a$}; % u node
  \node[latent, right=1cm of a] (y) {$y$}; % v node
  \node[latent, above=0.5cm of a, xshift=1cm] (x) {$x$}; % w node
  
  % Connect the nodes
  \edge{x}{a}; 
  \edge{x}{y};
  \edge{a}{y};
\end{tikzpicture}
\end{center}

The structural causal model is then
\begin{align}
    &x \leftarrow \epsilon_x \\
    &a \leftarrow f_a(x, \epsilon_a) \\
    &y \leftarrow f_y(a, x, \epsilon_y).
\end{align}

From this structural causal model, we can read out two important points. First, as we have learned earlier in \S\ref{sec:confounders-colliders-mediators}, $x$ is a confounder, and when it is not observed, the path $a \leftarrow x \to y$ is open, creating a spurious effect of $a$ on $y$. Second, the choice of $a$ is not fully determined by $x$. It is determined by the combination of $x$ and $\epsilon_a$, where the latter is independent of $x$. We are particularly interested in the second aspect here, since it gives us an opportunity to modify this graph by introducing a new variable that may help us remove the effect of the confounder $x$. 

We now consider an alternative to the structural causal model above by assuming that we found another variable $z$ that largely explains the exogenous factor $\epsilon_a$. That is, instead of saying that the action $a$ is determined by the combination of the covariate $x$ and an exogenous factor $\epsilon_a$, we now say that it is determined by the combination of the covariate $x$, this new variable $z$ and an exogenous factor $\epsilon'_a$. Because $z$ explains a part of the exogenous factor rather than $x$, $z$ is independent of $x$ {\it a priori}. This introduction of $z$ alters the structural causal model to become
\begin{align}
    &x \leftarrow \epsilon_x \\
    &z \leftarrow \epsilon_z \\
    &a \leftarrow f'_a(x, z, \epsilon'_a) \\
    &y \leftarrow f_y(a, x, \epsilon_y),
\end{align}
which corresponds to the following graph $G_1$:
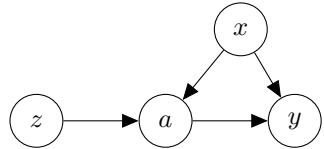
\begin{center}
\begin{tikzpicture}
  % Define nodes
  \node[latent] (z) {$z$}; % instrument variable
  \node[latent, right=1cm of z] (a) {$a$}; % u node
  \node[latent, right=1cm of a] (y) {$y$}; % v node
  \node[latent, above=0.5cm of a, xshift=1cm] (x) {$x$}; % w node
  
  % Connect the nodes
  \edge{z}{a};
  \edge{x}{a}; 
  \edge{x}{y};
  \edge{a}{y};
\end{tikzpicture}
\end{center}

This altered graph $G_1$ does not help us infer the causal effect of $a$ on $y$ any more than the original graph $G_0$ did. It however provides us with an opportunity to replace the original action $a$ with a proxy based purely on the newly introduced variable $z$ independent of $x$. 

We first notice that $x$ cannot be predicted from $z$, because $z$ and $x$ are by construction independent. The best we can do is thus to predict the associated action.\footnote{
    We can predict the marginal distribution over the action after marginalizing out $x$.
}
Let $p_{\tilde{a}}(a|z)$ be the conditional distribution induced by $g_{\tilde{a}}(z, \epsilon_a'')$. Then, 
we want to find $g_a$ that minimizes
\begin{align}
    \mathcal{E}_a = 
    -
    \mathbb{E}_{\epsilon_z} 
    \mathbb{E}_{\epsilon_x}
    \mathbb{E}_{\epsilon'_a} 
    \left[ 
        \log g_{\tilde{a}}(f'_a(\epsilon_x, z, \epsilon'_a), z)
    \right].
\end{align}

We also notice that $z$ cannot be predicted from $a$ perfectly without $x$, because $a$ is an observed collider, creating a dependency between $z$ and $x$. The best we can do is to thus to predict the expected value of $z$. 
That is, we look for $g_z$ that minimizes 
\begin{align}
    \mathcal{E}_z = 
    -
    \mathbb{E}_{\epsilon_z} 
    \mathbb{E}_{\epsilon_x}
    \left[
    \log p_{\tilde{z}}(\epsilon_z | 
    f'_a(\epsilon_x, \epsilon_z, \epsilon'_a))
    \right],
\end{align}
where we have used $p_{\tilde{z}}(z|a)$ be the conditional distribution induced by $g_{\tilde{z}}(a, \epsilon'_z)$,
similarly to $p_{\tilde{a}}(a|z)$ above.

Once we found reasonable solutions, $\hat{g}_{\tilde{a}}$ and $\hat{g}_{\tilde{z}}$, to the minimization problems above, respectively, we can further modify the structural causal model into
\begin{align}
    &a \leftarrow \hat{a} \\
    &x \leftarrow \epsilon_x \\
    &\tilde{z} \leftarrow \hat{g}_{\tilde{z}}(a, \epsilon'_z) \\
    &\tilde{a} \leftarrow \hat{g}_{\tilde{a}}(\tilde{z}, \epsilon''_a) \\
    &y \leftarrow f_y(\tilde{a}, x, \epsilon_y).
\end{align}
Since $x$ is really nothing but an exogenous factor of $y$ without impacting $\tilde{a}$ nor $z$ in this case, we can simplify this by merging $x$ and $\epsilon_y$ into
\begin{align}
    &a \leftarrow \hat{a} \\
    &\tilde{z} \leftarrow \hat{g}_z(a, \epsilon'_z) \\
    &\tilde{a} \leftarrow \hat{g}_{\tilde{a}}(\tilde{z}, \epsilon''_a) \\
    &y \leftarrow f_y(\tilde{a}, \epsilon'_y).
\end{align}
Because we assume the action $a$ is always given, we simply set it to a particular action $\hat{a}$. 

This structural causal model can be depicted as the following graph $G_2$:
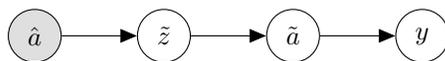
\begin{center}
\begin{tikzpicture}
  % Define nodes
  \node[obs] (a) {$\hat{a}$};
  \node[latent, right=1cm of a] (z) {$\tilde{z}$}; % instrument variable
  \node[latent, right=1cm of z] (a_recon) {$\tilde{a}$}; % u node
  \node[latent, right=1cm of a_recon] (y) {$y$}; % v node
  
  % Connect the nodes
  \edge{a}{z};
  \edge{z}{a_recon};
  \edge{a_recon}{y};
\end{tikzpicture}
\end{center}
In other words, we start from the action, approximately infer the extra variable $\tilde{z}$, approximately infer back the action and then predict the outcome. During two stages of inference ($\tilde{z} | a$ and $\tilde{a} | \tilde{z}$), we drop the dependence of $a$ on $z$. This happens, because $z$ was chosen to be independent of $x$ {\it a priori}.

In this graph, we only have two mediators in sequence, $\tilde{z}$ and $\tilde{a}$, from the action $a$ to the outcome $y$. We can then simply marginalize out both of these mediators in order to compute the interventional distribution over $y$ given $a$, as we learned in \S\ref{sec:confounders-colliders-mediators}. That is,
\begin{align}
    \hat{y}_{G_0}(a = \hat{a})
    = 
    \hat{y}_{G_1}(a = \hat{a})
    \approx
    \hat{y}_{G_2}(a = \hat{a})
    =
    \mathbb{E}_{\tilde{z}|\hat{a}}
    \mathbb{E}_{\tilde{a}|\tilde{z}}
    \mathbb{E}_{y|\tilde{a}}
    \left[
    y
    \right].
\end{align}

We call this estimator an {\it instrumental variable estimator} and call the extra variable $z$ an {\it instrument}. Unlike the earlier approaches such as regression and IPW, this approach is almost guaranteed to give you a biased estimate of the causal effect. 

Assume we are provided with $D=\left\{ (a_1, y_1, z_1), \ldots, (a_N, y_N, z_N) \right\}$ after we are done estimating those functions above. We can then get the approximate causal effect of the action of interest $\hat{a}$ by
\begin{align}
\label{eq:iv_estimate}
    \hat{y}_{\mathrm{IV}}(\hat{a}) 
    \approx
    \frac{
    \sum_{n=1}^N \mathds{1}(a_n = \hat{a})
    \mathbb{E}_{\epsilon''_a, \epsilon'_y}
    \hat{f}_y(\hat{g}_{\tilde{a}}(z_n, \epsilon''_a), \epsilon'_y)
    }{
    \sum_{n'=1}^N \mathds{1}(a_{n'} = \hat{a})
    }.
\end{align}
Additionally, if we are provided further with $D_{\bar{z}}=\left\{ (a'_1, y'_1), \ldots, (a'_{N'}, y'_{N'}) \right\}$, we can use $\hat{g}_z(a, \epsilon'_z)$ to approximate this quantity, together with $D$. Let $(a_n,y_n,r_n,z_n) \in \bar{D}$ be
\begin{align}
    (a_n,y_n,r_n,z_n)
    =
    \begin{cases}
        (a_n, y_n, 1, z_n),&\text{ if } n \leq N \\
        (a'_{n-N'+1}, y'_{n-N'+1}, 0, -1),&\text{ if } N < n \leq N+N'
    \end{cases}
\end{align}
Then, 
\begin{align}
    \hat{y}_{\mathrm{IV}}(\hat{a}) 
    \approx
    \sum_{n=1}^{N+N'}
    \frac{
    \mathds{1}(a_n = \hat{a})
    \mathbb{E}_{\epsilon''_a, \epsilon'_y, \epsilon'_z}
    \left[
    r_n
    \hat{f}_y(\hat{g}_{\tilde{a}}(
    z_n,
    \epsilon''_a), \epsilon'_y)
    +
    (1-r_n)
    \hat{f}_y(\hat{g}_{\tilde{a}}(
    \hat{g}_z(a_n, \epsilon'_z),
    \epsilon''_a), \epsilon'_y)
    \right]
    }{
    \sum_{n'=1}^{N+N'} \mathds{1}(a_{n'} = \hat{a})
    }.
\end{align}

In the former case, we must solve two regression problems, finding $\hat{f}_y$ and $\hat{g}_{\tilde{a}}$. When we do so by solving a least squares problem for each, we end up with two least squares problems that must be solved sequentially. Such a case is often referred as {\it two-stage least squares}. In the latter case, we benefit from extra data by solving three regression problems. Though, in most cases, we choose an easy-to-obtain instrument so that it is often enough to solve two regression problems.

There are two criteria that need to be considered when choosing an instrument $z$. First, the instrument must be independent of the confounder $x$ {\it a priori}. If this condition does not hold, we end up with the following graph:
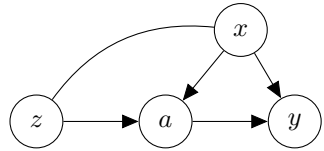
\begin{center}
\begin{tikzpicture}
  % Define nodes
  \node[latent] (z) {$z$}; % instrument variable
  \node[latent, right=1cm of z] (a) {$a$}; % u node
  \node[latent, right=1cm of a] (y) {$y$}; % v node
  \node[latent, above=0.5cm of a, xshift=1cm] (x) {$x$}; % w node
  
  % Connect the nodes
  \edge{z}{a};
  \edge{x}{a}; 
  \edge{x}{y};
  \edge{a}{y};
  \draw (z) to[bend left] (x);
\end{tikzpicture}
\end{center}

The undirected edge between $z$ and $x$ indicates that they are not independent. In this case, even if we manage to remove the edge between $a$ and $x$, there is still a spurious path $a \leftarrow z \leftrightarrow x \to y$, that prevents us from avoiding this bias. This first criterion is therefore the most important consideration behind choosing an instrument. 

\paragraph{Instrumental variables must be predictive of the action.}

The second criterion is that $z$ be a cause of $a$ together with $x$. That is, a part of whatever cannot be explained by $x$ in determining $a$ must be captured by $z$. We can see why this is important by recalling the IPW-based estimate against the instrumental variable based estimate. The IPW-based estimate from Eq.~\eqref{eq:ipw} is reproduced here as
\begin{align}
    \hat{y}_{\mathrm{IPW}}(\hat{a})
    \approx
    \frac{
    \sum_{n=1}^N
    \mathds{1}(a_n = \hat{a}) 
        \frac{y_n}
        {\hat{p}(\hat{a} | x_n)}
    }{
    \sum_{n'=1}^N
    \mathds{1}(a_{n'} = \hat{a})
    }.
\end{align}
If we contrast it with the instrument variable based estimate in Eq.~\eqref{eq:iv_estimate}, we get
\begin{align}
    \hat{y}_{\mathrm{IPW}}(\hat{a}) - \hat{y}_{\mathrm{IV}}(\hat{a}) 
    =
    \frac{
    \sum_{n=1}^N
    \mathds{1}(a_n = \hat{a}) 
    \overbrace{
    \left(
    \frac{y_n}
    {\hat{p}(\hat{a} | x_n)}
    -
    \mathbb{E}_{\epsilon_y, \epsilon''_a}
    \hat{f}_y(\hat{g}_{\tilde{a}}(z_n, \epsilon''_a), \epsilon_y)
    \right)
    }^{
    \mathrm{(a)}
    }
    }{
    \sum_{n'=1}^N
    \mathds{1}(a_{n'} = \hat{a})
    },
\end{align}
where we assume $D=\left\{ (a_1, y_1, x_1, z_1), \ldots, (a_N, y_N, x_N, z_N)\right\}$. 

There are two estimates within (a) above that result in a bias. Among these two, $\hat{f}_y$ and $\hat{g}_{\tilde{a}}$, the former does not stand a chance of being an unbiased estimate, because it is not given the unbiased estimate of the action nor the covariate $x$. Contrast this with the regression-based estimate above where $\hat{f}_y$ afforded to rely on the true (sampled) action and the true (sampled) covariate. The latter is however where we have a clear control over. 

Assume that $f^*_y$ is linear. That is,
\begin{align}
    f^*_y(a, x, \epsilon_y) = a^\top \alpha^* + x^\top \beta^* + \epsilon_y,
\end{align}
where $\alpha$ and $\beta$ are the coefficients. If $\mathbb{E}[x]=0$ and $a$ is selected independent of $x$, 
\begin{align}
    f^*_y(a, \epsilon_y) = a^\top \alpha^* + \epsilon_y.
\end{align}
The term (a), with the assumption that $\frac{y_n}{\hat{p}(a_n|x)} = f^*_y(a_n)$, can then be expressed as
\begin{align}
    \left(a_n - \mathbb{E}_{\epsilon''_a} \hat{g}_{\tilde{a}}(z_n, \epsilon''_a)\right)^\top \alpha^*
    - 
    \mathbb{E}_{\epsilon''_a}\left[\hat{g}_{\tilde{a}}(z_n, \epsilon''_a)\right]^\top r_{\alpha},
\end{align}
where $r_{\alpha}$ is the error in estimating $\alpha^*$, i.e., $\hat{\alpha}=\alpha^* + r_{\alpha}$.

For brevity, let $\hat{g}(z_n) = \mathbb{E}_{\epsilon''_a} \hat{g}_{\tilde{a}}(z_n, \epsilon''_a)$, which allows us to rewrite it as 
\begin{align}
    \left(a_n - \hat{g}(z_n)\right)^\top \alpha^* 
    -
    \hat{g}(z_n)^\top r_{\alpha}.
\end{align}

Let us now look at the overall squared error:
\begin{align}
    \frac{1}{M}
    \sum_{m=1}^M 
    \left(
    a_m^\top \alpha^* 
    -
    \hat{g}(z_m)^\top (\alpha^* - r_{\alpha})
    \right)^2,
\end{align}
where we use $m$ to refer to each example with $a_m=\hat{a}$. If we further expand the squared term,
\begin{align}
    &\frac{1}{M}
    \sum_{m=1}^M
    {\alpha^*}^\top a_m a_m^\top \alpha^* 
    +
    (\alpha^* - r_{\alpha})^\top
    \hat{g}(z_m) \hat{g}^\top(z_m)
    (\alpha^* - r_{\alpha})
    -2
    {\alpha^*}^\top 
    a_m \hat{g}^\top(z_m)
    (\alpha^* - r_{\alpha})
    \\
    &\approx
    {\alpha^*}^\top \mathbb{E}[a a^\top] \alpha^*
    +
    (\alpha^* - r_{\alpha})^\top \mathbb{E}[ \hat{g}(z) \hat{g}^\top(z) ](\alpha^* - r_{\alpha})
    - 
    2 
    {\alpha^*}^\top \mathbb{E}[a \hat{g}^\top(z)] (\alpha^* - r_{\alpha}).
\end{align}

The first term is constant. It simply tells us that the error would be greater if the variance of the relevant dimensions of the action on the outcome, where the relevancy is determined by $\alpha^*$, is great, the chance of mis-approximating it would be simply great as well. The second term tells us that the error would be proportional to the variance of the relevant dimensions of the predicted action on the outcome, where the relevancy is determined by the predicted coefficient, $\alpha^* - \hat{r}$. That is, if the variance of the predicted outcome is great, the chance of a large error is also great. The third term is where we consider the correlation between the true action and the predicted action, again along the dimensions of relevance. 

This derivation tells us that the instrument must be selected to be highly predictive of the action (the third term) but also exhibit a low variance in its prediction (the second term). Here comes the classical dilemma of bias-variance trade-off in machine learning.

\paragraph{The linear case.}

The instrument variable approach is quite confusing. Consider a 1-dimensional fully linear case here in order to build up our intuition. Assume
\begin{align}
    &x \leftarrow \epsilon_x \\
    &a \leftarrow \gamma x + \epsilon_a \\
    &y \leftarrow \alpha a + \beta x + \epsilon_y,
\end{align}
where $\epsilon_x$ and $\epsilon_y$ are both zero-mean Normal variables. If we intervene on $a$, we would find that the expected outcome equals
\begin{align}
    \mathbb{E}[y|\mathrm{do}(a)] = \alpha a,
\end{align}
and thereby the ATE is
\begin{align}
    \mathrm{ATE} = \mathbb{E}[y|\mathrm{do}(a=1)] - \mathbb{E}[y|\mathrm{do}(a=0)] = \alpha.
\end{align}

With a properly selected instrument $z$, that is, $z \indep x$, we get
\begin{align}
    &z \leftarrow \epsilon_z \\
    &a \leftarrow \gamma x + \psi z + \epsilon'_a. 
\label{eq:instrument-predicted-action}
\end{align}
Because $z\indep x$, the best we can do is to estimate $\psi$ to minimize
\begin{align}
\label{eq:iv-1st-ls}
    \min_\psi \sum_{n=1}^N (a_n - \psi z_n)^2,
\end{align}
given $N$ $(a,z)$ pairs. 
The minimum attainable loss is 
\begin{align}
    (\gamma x)^2,
\end{align}
assuming zero-mean $\epsilon'_a$, because the contribution from $x$ cannot be explained by the instrument $z$. 

With the estimated $\hat{\psi}$, we get 
\begin{align}
    \hat{a} \leftarrow \hat{\psi} z + \epsilon''_a,
\end{align}
and know that
\begin{align}
    \hat{a} = a - \gamma x
\end{align}
on expectation. 

By plugging in $\hat{a}$ into the original structural causal model, we end up with
\begin{align}
    &x \leftarrow \epsilon_x \\
    &\hat{a} \leftarrow \hat{\psi} z + \epsilon''_a 
    \label{eq:iv-hat-a}
    \\
    &y \leftarrow \tilde{\alpha} \hat{a} + \beta x + \epsilon_y'.
\end{align}

We can now estimate $\tilde{\alpha}$ by minimizing
\begin{align}
\label{eq:iv-2nd-ls}
    \sum_{n=1}^N
    \left(
    y_n - \tilde{\alpha} \hat{a}_n
    \right)^2,
\end{align}
assuming both $\epsilon_x$ and $\epsilon_y'$ are centered. 

If we assume we have $x$ as well, we get
\begin{align}
    &
    \sum_{n=1}^N
    \left(
    y_n - \tilde{\alpha} (a_n - \gamma x_n)
    \right)^2 
    \\
    &=
    \sum_{n=1}^N
    \left(
    y_n - \alpha (a_n - \gamma x_n)
    + (\alpha - \tilde{\alpha}) (a_n - \gamma x_n)
    \right)^2 
    \\
    &=
    \sum_{n=1}^N
    \left(
    \epsilon_{y,n}
    + (\alpha - \tilde{\alpha}) \epsilon_{a,n}'
    \right)^2 
    \\
    &=
    \sum_{n=1}^N
    \left(
    \epsilon_{y,n}^2
    +
    (\alpha - \tilde{\alpha})^2 (\epsilon_{a,n}')^2
    + 2 (\alpha - \tilde{\alpha}) \epsilon_{y,n} \epsilon_{a,n}'
    \right)
    \\
    &=_{n\to\infty}
    \mathbb{V}[\epsilon_y]
    +
    (\alpha - \tilde{\alpha})^2
    \mathbb{V}[\epsilon_a'].
\end{align}
The first term is irreducible, and therefore we focus on the second term. The second term is the product of two things. The first one, $(\alpha - \tilde{\alpha})^2$, measures the difference between the correct $\alpha$ and the estimated effect. It tells us that minimizing this loss w.r.t. $\tilde{\alpha}$ is the right way to approximate the true causal effect $\alpha$. 
% This term is scaled by the variance of $\epsilon_a'$ which represents how much variance of the action $a$ remains after predicting $a$ using the instrument $z$, as from Eq.~\eqref{eq:instrument-predicted-action}.

We can plug in $\hat{a}$ from Eq.~\eqref{eq:iv-hat-a} instead:
\begin{align}
    &
    \sum_{n=1}^N
    \left(
    y_n - \tilde{\alpha} (\hat{\psi} z_n + \epsilon''_{a,n})
    \right)^2 
    \\
    &=
    \sum_{n=1}^N
    \left(
    (y_n - \tilde{\alpha} \hat{\psi} z_n) + \tilde{\alpha} \epsilon''_{a,n}
    \right)^2 
    \\
    &=
    \sum_{n=1}^N
    \left(
    (y_n - \tilde{\alpha} \hat{\psi} z_n)^2
    +
    \tilde{\alpha}^2 (\epsilon''_{a,n})^2
    +
    2 (y_n - \tilde{\alpha} \hat{\psi} z_n) \tilde{\alpha} \epsilon_{a,n}''
    \right)
    \\
    &=_{n\to\infty}
    \sum_{n=1}^N
    (y_n - \tilde{\alpha} \hat{\psi} z_n)^2
    +
    \tilde{\alpha}^2 \mathbb{V}[\epsilon''_a].
\end{align}
The first term is about how predictive the instrument $z$ is of $y$, which is a key consideration in choosing the instrument. If the instrument is not predictive of $y$, the instrument variable approach fails dramatically. The second term corresponds to the variance of the action not explained by the instrument, implying that the instrument must also be highly correlated with the action. 

In this procedure, we have solved least squares twice, \eqref{eq:iv-1st-ls} and \eqref{eq:iv-2nd-ls}, which is a widely used practice with instrument variables. We also saw the importance of the choice of the instrument variable.

\paragraph{An example: taxation}

One of the most typical example of an instrument is taxation. It is particularly in the United States of America (USA), due to the existence of different tax laws and rates across fifty states. For instance, imagine an example where the action is cigarette smoking, the outcome is the contraction of lung cancer and the confounder is an unknown genetic mutation that both affects the affinity to nicotine addiction and the incidence of a lung cancer. Because we do not know such a genetic mutation, we cannot easily draw a conclusion about the causal effect of cigarette smoking on lung cancer. There may be a spurious correlation arising from this unknown, and thereby unobserved, genetic mutation. Furthermore, it is definitely unethical to randomly force people to smoke cigarettes, which prevents us from running an RCT.

We can instead use state-level taxation on tobacco as an instrument, assuming that lower tax on tobacco products would lead to a higher chance and also rate of smoking, and vice versa. First, we build a predictor of smoking from the state (or even county, if applicable) tax rate. The predicted amount of cigarettes smoked by a participant can now work as a proxy to the original action, that is the actual amount of cigarettes smoked. We then build a predictor of the incidence of lung cancer as well as the reverse predictor (action-to-instrument prediction). We can then use one of the two instrument variable estimators above to approximate the potential outcome of smoking on lung cancer. 

\section{Summary}

In this chapter, we have learned the following concepts:

\begin{enumerate}
    \item Challenges in active causal inference: practical, ethical and legal challenges
    \item When confounders were observed: Regression, inverse probability weighting and matching
    \item When confounders were not observed: instrument variables
\end{enumerate}

There are a few other widely used passive causal inference algorithms, but they are left for the final section on \S\ref{chap:remaining-topics} Remaining Topics, such as difference-in-difference, regression discontinuity and double machine learning.

\chapter{Causality and Machine Learning}

In this chapter, we finally delve in to the `machine learning' side of this course, which is titled `Introduction to Causal Inference in Machine Learning'. In order to do so, we need to start by establishing when we do not need to think of causal inference, or more broadly causality, in machine learning. After establishing it, we will move on to the other extreme, where conventional machine learning cannot do anything on its own. We then try to incorporate some of the concepts we have learned so far, in order to land between these two extreme cases and solve some of the most challenging and important problems in modern machine learning.

\section{Out-of-Distribution Generalization}

\subsection{Setup: I.I.D.}

We must start by defining what we mean by `prediction'. In this particular course, we first assume that each and every input-output pair $(x,y)$, input $x$ or output $y$ is sampled independently of each other. This is a pretty strong assumption, since the world often changes based on what we have seen, because those who saw a sample pair may and often do change their behaviors. For instance, consider building a stock price forecasting model. Once you use a predictor to predict whether the price of a particular stock goes up or down and trade based on the outcome, the next input $x$, that is the stock of your next interest, is not anymore independently selected but based on your own success/failure from the previous trade. 

This assumption is however also reasonable, because there are many phenomena in which our behaviours do not matter much in a reasonably short horizon. For instance, consider installing and using a bird classifier at a particular forest. With a fixed camera, the input to this classifier will be largely independent of which birds (or not) were seen earlier, although 
spotting of a particular bird may attract poachers to this forest who would dramatically affect the bird population in a longer time frame. 

Next, we assume that all these pairs are drawn from the `identical' distribution. This is similar if not identical to the stationarity assumption from RCT. In RCT, we often rely on a double blind experiment design, in order to ensure that the causal effect $p^*(y|a,x)$ does not change over the trial. In this section as well as conventional statistical learning theory, we assume all input-output pairs were drawn from the same distribution. 

Combining these two assumptions, we arrive at a so-called training set $D$ which satisfies
\begin{align}
    p(D) = \prod_{(x,y) \in D} p^*(x, y),
\end{align}
according to the definition of independence. 
We do not have access to nor have knowledge of $p^*$. 
We use this training set $D$ for both model fitting (training) and selection (validation).

Once the predictive model $\hat{p}$ is ready, we deploy it to make a prediction on a novel input $x'$ drawn from a distribution $q^*$. That is,
\begin{align}
    \hat{y} \sim \hat{p}(y | x'),
\end{align}
where $(x', y') \sim q^*$. We are often not given $y'$. After all, $y'$ is what we want to use our predictive model to infer. 

We say that the predictive model is accurate, if the following quantity is low:
\begin{align}
    R(\hat{p})
    =
    \mathbb{E}_{(x',y') \sim q^*} 
    \left[
    l(y', \hat{p}(y|x'))
    \right],
\end{align}
where $l(\cdot, \cdot) \geq 0$ is the loss (misclassification rate). 

In traditional statistical learning theory, $q^*$ is assumed to be $p^*$, and under this assumption, the goal of designing a learning algorithm is to minimize a so-called excess risk:
\begin{align}
    R_{\mathrm{excess}}(\hat{p}) = R(\hat{p}) - R(p^*)
\end{align}
with respect to $\hat{p}$. 
Since we do not have access to $p^*$, we often use Monte Carlo approximation to compute $R(\hat{p})$, as follows
\begin{align}
\label{eq:empirical-risk}
    R(\hat{p})
    \approx
    \hat{R}_N(\hat{p})
    = 
    \frac{1}{N} 
    \sum_{n=1}^N 
    l(y_n, p(y|x_n)),
\end{align}
where $(x_n, y_n) \sim p^*$. 

With a (strong) assumption of uniform convergence, which is defined as 
\begin{align}
    \sup_{\hat{p}} \left| R(\hat{p}) - \hat{R}_N(\hat{p}) \right| \to_p 0,
\end{align}
we can minimize $R$ using $\hat{R}$ with a large enough data set, i.e., $N \to \infty$, and find a good predictive model $\hat{p}$. Of course, since $N$ is always finite in reality, there is almost always non-zero generalization error. 

Since we never have access to $R(\hat{p})$ even after learning, it is a usual practice to use a separate (held-out) set of examples again drawn from the same distribution $p^* = q^*$ as the test set to approximate the generalization error of a trained model $\hat{p}$. Let $D' = \left\{ (x'_1,y'_1), \ldots, (x'_K, y'_K) \right\}$. Then,
\begin{align}
    R(\hat{p}) \approx 
    \frac{1}{K}
    \sum_{k=1}^K
    l(y'_k, \hat{p}(y|x'_k)).
\end{align}
Such a test-set accuracy, or more simply a test accuracy, has been a workhorse behind rapid advances in machine learning over the past several decades. 

With this whole paradigm in your mind, it is important to notice that the key assumption here is $q^*(x,y)=p^*(x,y)$. In other words, we assume that an instance a predictive model would be tested in the deployment would follow the same distribution as that from which the training examples were drawn, i.e., $q^*(x) = p^*(x)$. Furthermore, the conditional distribution over the outcome does not change either, i.e., $q^*(y|x) = p^*(y|x)$. In this case, there is no reason for us to consider the underlying generating process behind $p^*$ nor $q^*$ separately. 

\subsection{Out-of-Distribution Generalization}

\paragraph{Impossibility of Out-of-Distribution (ood) generalization.}

In reality, it is rarely that $q^* = p^*$, because the world changes. When $q^* \neq p^*$, we must be careful about discussing generalization. We must be careful, because we can always choose $q^*$ to be such that minimizing $R(\hat{p})$ in Eq.~\eqref{eq:empirical-risk} would lead to maximizing
\begin{align}
    R^{q^*}(\hat{p})
    =
    \mathbb{E}_{(x,y) \sim q^*}[
    l(y, \hat{p}(y|x))
    ].
\end{align}

Assume $y \in \left\{0, 1\right\}$. Consider the following $q^*$, given $p^*(x,y) = p^*(x) p^*(y|x)$,
\begin{align}
    q^*(x, y) = p^*(x) q^*(y|x),
\end{align}
where
\begin{align}
\label{eq:contrarian-q}
    q^*(y|x) 
    = 1 - p^*(y|x).
\end{align}
That is, the mapping from $x$ to $y$ is reversed. When $x$ was more probable to be observed together with $y=1$ under $p^*$, it is now more probable to be observed together with $y=0$ now under $q^*$, and vice versa. 

If we take the log loss, which is defined as
\begin{align}
    l(y, \hat{p}(y|x)) = -\log \hat{p}(y|x),
\end{align}
learning corresponds to minimizing the KL divergence from the true distribution to the learned, predictive distribution. Mathematically,
\begin{align}
    \arg\min_{\hat{p}} \frac{1}{N} \sum_{n=1}^N l(y_n, \hat{p}(y_n|x_n))
    \approx
    \arg\min_{\hat{p}} \mathbb{E}_x \mathrm{KL}(
    p^*(\cdot | x) 
    \|
    \hat{p}(\cdot |x)
    ).
\end{align}
In other words, learning corresponds to recovering $p^*$ as much as we can for as many probable $x$'s under $p^*(x)$. 

It is clear that minimizing this loss function would make our predictive model worse on a new distribution \eqref{eq:contrarian-q}. Because the following holds for any particular example $(x,y)$:
\begin{align}
    \log p^*(y|x) = \log (1 - q^*(y|x)).
\end{align}
Since $\log$ is a monotonic function, maximizing $p^*$ is equivalent to minimizing $q^*$. As soon as we start minimizing the log loss for learning, out-of-distribution generalization to $q^*$ gets worse, and there is no way to avoid it, other than not learning at all.

This is a simple but clear example showing how out-of-distribution generalization is not possible in general. There will always be a target distribution that disagrees with the original distribution, such that learning on the latter is guaranteed to hurt the generalization accuracy on the former. In general, such a target distribution can be written down as
\begin{align}
    \log q^*(y|x) \propto \log (1 - p^*(y|x)). 
\end{align}
We can also come up with a similar formula for $q^*(x)$, such that there is almost no support overlap between $p^*(x)$ and $q^*(x)$.

\paragraph{Out-of-distribution generalization.}

We then must narrow down the scope in order to discuss out-of-distribution generalization. There are many different ways to narrow the scope, and one way is to ensure that the target distribution $q^*$ is not too far from the original distribution $p^*$. Let $D: \mathcal{P} \times \mathcal{P} \to \mathrm{R}_+$ be a (asymmetric) divergence between two distributions, such that the larger $D(p, q)$ implies the greater difference between these two distributions, $p$ and $q$. Then, we can write a so-called distributionally-robust loss as
\begin{align}
    \min_{\hat{p}}
    \sup_{q: D(p^*,q)\leq \delta}
    \mathbb{E}_{(x,y) \sim q} \left[ 
    l(y, \hat{p}(y|x))
    \right],
\end{align}
where $\sup$ is the supremum which is the smallest item that is greater than equal to all the other items in a partially ordered set~\citep{shapiro2017distributionally}. 

The distributionally-robust loss above minimizes ($\min_{\hat{p}}$) the expected loss ($\mathbb{E}_{(x,y) \sim q} \left[ l(y, \hat{p}(y|x)) \right]$) over the worst-case distribution ($\sup_{q}$) within the divergence constraint ($q: D(p^*,q)\leq \delta$). Despite its generality, due to the freedom in the choice of the divergence $D$ and the universality (the worst case), such distributionally-robust optimization is challenging to use in practice. The challenge mainly comes from the fact that we must solve a nested optimization problem, where for each update of $\hat{p}$ we must solve another optimization problem that maximizes the loss w.r.t. the distribution $q$. This problem can be cast as a two-player minimax game which is more challenging, both in terms of convergence and its speed, than a more conventional optimization problem.
Furthermore, it is often unclear how to choose an appropriate divergence $D$ and the threshold $\delta$, as these choices are not grounded in the problem of interest.

Instead, we are more interested in an alternative to the distributionally robust optimization approach. Instead of specifying a divergence, we can describe how the distribution changes in terms of the probabilistic graphical model, or equivalently the structural causal model underlying $p^*$ and $q^*$. Depending on such a distributional change, we may be able to characterize the degree of generalization or even to come up with a better learning algorithm. 

\subsection{Case Studies}

\paragraph{The label proportion shift.}

Let us consider a very basic example of a generative classier which assumes the following generating process:

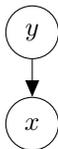
\begin{center}
\vspace{5mm}
\begin{tikzpicture}
  % Define nodes
  \node[latent] (y) {$y$}; % u node
  \node[latent, below=0.5cm of y] (x) {$x$}; % v node
  
  % Connect the nodes
  \edge{y}{x}; % Edge from u to v 
\end{tikzpicture}
\vspace{5mm}
\end{center}

Under this generating process, the joint probability is written as
\begin{align}
    p^*(x,y) = p^*(y) p^*(x|y),
\end{align}
and the posterior distribution over the output $y$ is
\begin{align}
    p(y|x) = \frac{p(y) p(x|y)}{p(x)} = \frac{p(y) p(x|y)}{\sum_{y' \in \mathcal{Y}} p(y') p(x|y')}.
\end{align}

Given a training set $D=\left\{ (x_1, y_1), \ldots, (x_N, y_N) \right\}$, where each $(x_n,y_n)$ was drawn from the generating process above, that is,
\begin{align}
    &y_n \sim p^*(y) \\
    &x_n \sim p^*(x|y_n).
\end{align}
We can train a neural network classifier that takes as input $x$ and outputs a probability for each possible value of $y$. 
This neural network can be written as
\begin{align}
\label{eq:softmax-nn}
    \hat{p}(y|x; \theta, b) = 
    \frac{\exp(f_y(x; \theta) + b_y)}
    {\sum_{y' \in \mathcal{Y}} \exp(f_{y'}(x; \theta)+ b_{y'})},
\end{align}
where $f_y(x; \theta)$ is the $y$-th element of the $|\mathcal{Y}|$-dimensional output from the neural network $f$, parametrized by $\theta$ and the bias vector $b \in \mathbb{R}^{|\mathcal{Y}|}$. 

Inspecting this neural net's formulation, based on the so-called softmax output, we notice the following correspondences:
\begin{enumerate}
    \item $p^*(y) \approx \frac{1}{Z_y} \exp(b_y)$
    \item $p^*(x|y) \approx \frac{1}{Z_{x|y}} \exp(f_y(x; \theta))$,
\end{enumerate}
where $Z_y$'s and $Z_{x|y}$'s are the normalization constants, which are cancelled out in Eq.~\eqref{eq:softmax-nn}.\footnote{
$\exp(a + b) = \exp(a) \exp(b)$.
} 
In other words, the bias $b_y$ captures the marginal distribution over the output, and the rest the conditional distribution over the input given the output. 

This view suggests a two-stage learning process. In the first stage, we simply set $b_y$ to be $\log p^*(y)$ (and thereby set $Z_y=1$ implicitly.) Then, we use optimization, such as stochastic gradient descent, to estimate the rest of the parameters, $\theta$. After learning is over, we get
\begin{align}
    \label{eq:y-given-x}
    \hat{p}(y|x) = \hat{p}(y) \frac{\exp(f_y(x; \hat{\theta}))}{\sum_{y'} \exp(f_{y'}(x; \hat{\theta}))}.
\end{align}
It is important to notice that the second term on the right hand side is not the estimate of $p^*(x|y)$, since the denominator must include the extra normalization, i.e. $p(x)$. In other words, 
\begin{align}
    \frac{\exp(f_y(x; \hat{\theta}))}{\sum_{y'} \exp(f_{y'}(x; \hat{\theta}))} = 
    \frac{\hat{p}(x|y)}{\hat{p}(x)}.
\end{align}

This predictive model $\hat{p}(y|x)$ would work well even on a new instance under the iid assumption, that is, $p^*(y|x)=q^*(y|x)$. It is however not the case, because $q^*(y) \neq p^*(y)$. For instance, imagine we trained a COVID-19 diagnosis model based on various symptoms, including cough sound, temperature and others, during the winter of 2021. During this period, COVID-19 was rampant, that is, $p^*(y=1)$ was very high. If we use this model however in the winter of 2024, the overall incident rate of COVID-19 is much lower. In other words, $q^*(y=1) \ll p^*(y=1)$. This would lead to the overestimation of $p(y=1|x)$, because the prediction is proportional to $\hat{p}(y=1)$ which is an estimate of the outdated prior $p^*(y=1)$ over the output not of the latest prior $q^*(y=1)$. The prediction becomes worse as $q^*$ deviates further away from $p^*$. 

One simple way to address this is to assume that {\it a priori} it is more probable for the label marginal, i.e., the marginal distribution over the output, to be closer to the uniform distribution. This is a reasonable assumption in many contexts when we are not allowed any information about the situation. For instance, it is perfectly sensible to assume that any given coin is likely to be fair (that is, it has the equal chance of landing head or tail.) In that case, we would simply set the bias $b$ to be an all-zero vector so that
\begin{align}
    \hat{p}(y|x) = \frac{\exp(f_y(x; \hat{\theta}))}{\sum_{y'} \exp(f_{y'}(x; \hat{\theta}))}.
\end{align}

Sometimes we are given some glimpse into $q^*$. In the case of COVID-19, it is difficult to collect $(x,y)$ pairs but it is often easy to collect $y$'s by various means, including the survey and rapid testing in various event venues. Let $\hat{q}(y)$ be the estimate of $q^*(y)$ from such a source. We can then replace $\hat{p}(y)$ with this new estimate in Eq.~\eqref{eq:y-given-x}, resulting in
\begin{align}
    \hat{p}(y|x) = \hat{q}(y) \frac{\exp(f_y(x; \hat{\theta}))}{\sum_{y'} \exp(f_{y'}(x; \hat{\theta}))}.
\end{align}
This is equivalently to replacing the bias $b_y$ with $\log \hat{q}(y)$. 

In practice, it is often the case that the number of $y$ samples we can collect is limited, leading to a high-variance estimate of $q^*$. We do not want to rely solely on such an estimate. Instead, we can interpolate between $\hat{p}(y)$ and $\hat{q}(y)$, leading to replacing the bias of each output with
\begin{align}
    b_y \leftarrow \log \left(\alpha \hat{p}(y) + (1-\alpha) \hat{q}(y) \right),
\end{align}
with $\alpha \in [0, 1]$. $\alpha$ describes the degree of our trust in the original estimate of the label marginal. if $\alpha = 1$, we end up with the original iid setup, and with $\alpha=0$, we fully trust our new estimate of the label marginal. 

\paragraph{Data augmentation.}

Consider an object classification task, where the goal is to build a classifier that categorizes the object in the center of an image into one of $K$ predefined classes. Just like before, we assume generative classification in which the object label produces the image. We however further assume that there exists an extra variable $z=(i,j)$ that determines the precise position of the object. 

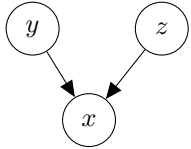
\begin{center}
\vspace{5mm}
\begin{tikzpicture}
  % Define nodes
  \node[latent] (y) {$y$}; % u node
  \node[latent, right=1cm of y] (z) {$z$}; % u node
  \node[latent, below=0.5cm of y, xshift=0.75cm] (x) {$x$}; % v node
  
  % Connect the nodes
  \edge{y}{x}; % Edge from u to v 
  \edge{z}{x};
\end{tikzpicture}
\vspace{5mm}
\end{center}

During the training time, $z$ follows a Normal distribution centered at the center of the image, i.e., $z \sim \mathcal{N}(\mu_z=[0, 0]^\top, I_2)$. Assuming that the background is randomly produced and does not correlate with the identity of the object in the center, a classifier we train on data produced from this data generating process should become blind to periphery pixels, since $\mathrm{cov}(x_{mn}, y) \approx 0$, where $|m| \gg 0$ and $|n| \gg 0$. This can be written down as
\begin{align}
    p(x_{mn} | y) \approx p(x_{nm}),
\end{align}
meaning that $x_{mn}$ is independent of $y$. 

If we make the na\"ive Bayes assumption, that is, all pixels are independent conditioned on the label, we get the following expression of the posterior over the label:
\begin{align}
    p(y|x) \propto p(y) \prod_{m, n} p(x_{mn} | y) 
    \propto
    p(y) \prod_{(m, n) \in C} p(x_{mn} | y), 
\end{align}
where $C$ is a set of pixels near the center. In other words, if the object is outside the center of the image, the posterior distribution over the label would not capture the actual identity of the object. 

This dependence on the position arises from the existence of the hidden variable $z$ {\it and} its prior distribution $p^*(z)$. If this prior distribution over $z$ shifts in the test time, such that $q^*(z) = \mathcal{N}(\mu_z=[100, 100]^\top, I_2)$, all objects in the images would be positioned on the top-right corners. The classifier based on the training set with $p^*(z)$ will then completely fail to detect and classify these objects. 

Because we assume to know the precise type of shift that is possible, we can now mitigate this issue by data augmentation~\citep{yaeger1996effective}. During training, we randomly shift a training image such that the position of the object in the image varies more greatly than it usually does in the original training set. This can be thought of as introducing another random variable $u$ such that 
\begin{align}
    p(l | z, u) = p(l),
\end{align}
where $l$ indicates the position of the object in an image. In other words, $u$ makes the position of an object independent of $z$, such that a classifier trained on the training data with such data augmentation is able to detect objects in any position, making it invariant to the distributional shift of $z$.

\section{Invariance: Stable Correlations are Causal Correlations}

Once we have a probabilistic graphical model, or a structural causal model, that describes the generating process and have a crisp idea of which distribution shifts how, we can come up with a learning algorithm that may alleviate the detrimental effect of such a distribution shift. It is however rare that we can write down the description of a generating process in detail. It is even rarer to have a crisp sense of how distributions shift between training and test times. For instance, how would you describe relationships among millions of pixels of a photo and unobserved identities of objects within it? 

We can instead focus on devising an alternative way to determine which correlations are considered causal and which other correlations are spurious. The original way to distinguish causal and spurious correlations was entirely reliant on the availability of a full generating process in the form of a probabilistic graphical model. One alternative is to designate correlation that holds both during training and test time as causal, and the rest as spurious~\citep{peters2016causal}. In other words, any correlation that is {\it invariant} to the distributional shift is considered causal, while any correlation that varies according to the distributional shift is considered spurious. The goal is then to find a learning algorithm that can ignore spurious (unstable) correlations while capturing only (stable) causal correlations, for the purpose of prediction. 

\subsection{An Environment as a Collider}

\paragraph{A case study: a bird or a branch?}

Imagine a picture of a bird taken from a forest. The bird is probably somewhere near the center of the photo, since the bird is the object of interest. It is extremely difficult to take a good picture of a flying bird, and hence, it is highly likely that the bird is not flying but is sitting. Since we are in a forest, it is highly likely that the bird is sitting on a tree branch with the branch placed near the bottom of the photo. Compare this to a picture with a bird taken from the same forest. the chance of a tree branch being solely near the bottom of the photo is pretty slim. After all, it is a forest, and there are many branches all over. I can then create a bird detector using either of two features; one is a feature describing a bird near the center and the other is a feature describing the location of a tree branch. Clearly, we want our bird detector to use the first feature, that is, to check whether there is a bird in the picture rather than whether there is a tree branch near the bottom of the picture, in order to tell whether there is a bird in the picture. Either way, however, the bird detector would work pretty well in this situation.

A bird detector that relies on the position of a tree branch would not work well if suddenly all the pictures are from indoors rather than from a forest. Most of the birds indoors would be confined in their cages and would not be sitting on tree branches. Rather, they would be sitting on an artificial beam or on the ground. On the other hand, a bird detector that relies on the actual appearance features of a bird would continue to work well. That is, the correlation between the label (`bird' or not) and the position of a tree branch (`bottom' or not) is not stable, while the correlation between the label and the bird-like appearance of a bird is stable. That is, the former is spurious, while the latter is causal. A desirable bird detector would rely on the causal correlation and discard any spurious correlation during learning. 

\paragraph{An environment indicator is a collider.}

A precise mechanism by which these unstable correlations arise can be extremely complex and is often unknown. In other words, we cannot rely on having a precise structural casual model from which we can read out all paths between the input and output, designate each as causal or spurious and adjust for those spurious paths. Instead, we can think of an extremely simplified causal model that includes only three variables; input $x$, output $y$ and collider $z$, as in

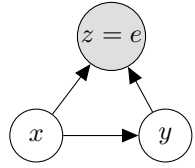
\begin{center}
\vspace{5mm}
\begin{tikzpicture}
  % Define nodes
  \node[latent] (u) {$x$}; % u node
  \node[latent, right=1cm of u] (v) {$y$}; % v node
  \node[obs, above=0.5cm of u, xshift=1cm] (w) {$z=e$}; % w node
  
  % Connect the nodes
  \edge{u}{v}; % Edge from u to v 
  \edge{u}{w};
  \edge{v}{w};
\end{tikzpicture}
\vspace{5mm}
\end{center}

In this causal model, the collider $z$ tell us whether we are in a particular environment (e.g. a forest above.) When we collect data from this causal model while being conditioned on a particular environment, this conditioning on the collider opens the path $x \to z \leftarrow y$, as we have learned earlier in \S\ref{sec:confounders-colliders-mediators}.

This way of thinking necessitates a bit of mental contortion. Rather than saying that a particular environment affects the input and output, but we are saying that a particular combination of the input and output probabilistically defines an environment. That is, $p(z | x, y)$ is the distribution defined over all possible environments $z$ given the combination of $x$ and $y$. Indeed, if $x$ is a picture with a tree branch near the bottom of a picture and $y$ states that there is a bird, the probability of $z$ being a forest is quite high. The environment dependence can then be thought of as drawing training instances from the graph above where the environment $w$ takes a particular target environment value (e.g. `forest'.) 

The most naive solution to this issue is to collect as much extra data as possible while avoiding such `selection bias' arising from conditioning the collider $z$ on any particular value. If we do so, it is as if the collider $z$ did not exist at all, since marginalizing out $z$ leads to the following simplified graph:

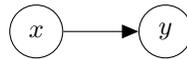
\begin{center}
\vspace{5mm}
\begin{tikzpicture}
  % Define nodes
  \node[latent] (u) {$x$}; % u node
  \node[latent, right=1cm of u] (v) {$y$}; % v node
  
  % Connect the nodes
  \edge{u}{v}; % Edge from u to v 
\end{tikzpicture}
\vspace{5mm}
\end{center}

A predictive model fitted on this graph $\hat{p}(y|x)$ would capture the causal relationship between the input and output, since the conditional and interventional distributions coincide in this case, that is, $p^*(y|x) = p^*(y|\mathrm{do}(x))$. This approach is however often unrealistic. 

\subsection{The Principle of Invariance}
\label{sec:invariance}

\paragraph{Invariant features.}

So far, we have considered each variable as an unbreakable unit. This is however a very strong assumption, and we should be able to easily split any variable into two or more pieces. This is in fact precisely what we often do by representing an object as a $d$-dimensional vector by embedding it into the $d$-dimensional Euclidean space. We are splitting a variable $x$ into a set of $d$ scalars which collectively representing the value the variable takes. We can then look at a subset of these dimensions and instead of the full variable, in which case the statistical as well as causal relationships with other variables may change. This applies even to a 1-dimensional random variable, where we can apply a nonlinear function to alter its relationship with other variables. 

Consider the following structural causal model:
\begin{align}
    &x \leftarrow \epsilon_x, \\
    &z \leftarrow \mathds{1}(x > 0) \max(0, x + \epsilon_z), \\
    &y \leftarrow \mathds{1}(x \leq 0) \min(0, x + \epsilon_y) + z , 
\end{align}
where
\begin{align}
    &\epsilon_x \sim \mathcal{N}(0, 1^2) \\
    &\epsilon_z \sim \mathcal{N}(0, 1^2) \\
    &\epsilon_y \sim \mathcal{N}(0, 1^2).
\end{align}
this model simplifies to $y \sim \mathcal{N}(0, 1^2 + 1^2)$, where two unit variances come from $\epsilon_x$ and either $\epsilon_z$ or $\epsilon_y$ depending on the sign of $x$. With the following nonlinear function applied to $x$, however, $y$ takes a different form:
\begin{align}
    g(x) = \mathds{1}(x \leq 0) x.
\end{align}
By replacing $x$ with $g(x)$ above,
\begin{align}
    p(y) 
    \propto
    \begin{cases}
        0,&\text{ if } y > 0, \\
        \mathcal{N}(y; 0, 1^2 + 1^2),&\text{ otherwise} \\
    \end{cases}
\end{align}
This has the effect of removing the correlation flowing through the path $x \to z \to y$, leaving only $x \to y$, because $z$ is now a constant function regardless of the value $x$ takes. By inspecting the relationship between $g(x)$ and $y$, we can measure the direct causal effect of $x$ on $y$. 

This example illustrates that there may be a nonlinear function of $x$ that may results in a variable that preserves enough information to prepare the direct causal relationship between $x$ and the output $y$ but removes any relationship $x$ has with the other variables in the structural causal model. In the context of the environment variable $z$, which is a collider, the goal is then to find a feature extractor $g$ such that the original graph is modified into

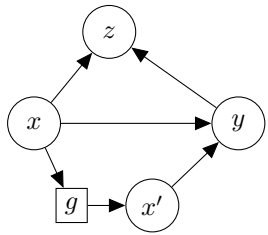
\begin{center}
\vspace{5mm}
\begin{tikzpicture}
  % Define nodes
  \node[latent] (x) {$x$}; % u node
  \node[draw, rectangle, below=0.5cm of x, xshift=0.5cm] (g) {$g$};
  \node[latent, right=0.5cm of g] (gx) {$x'$};
  \node[latent, right=2cm of x] (y) {$y$}; % v node
  \node[latent, above=0.5cm of x, xshift=1cm] (z) {$z$}; % w node
  
  % Connect the nodes
  \edge{x}{g}; % Edge from u to v 
  \edge{g}{gx};
  \edge{gx}{y};
  \edge{x}{z};
  \edge{y}{z};
  \edge{x}{y};
\end{tikzpicture}
\vspace{5mm}
\end{center}

Ideally, we want $g$ such that $g(x)$ explains the whole of $x$'s direct effect on $y$. That is,
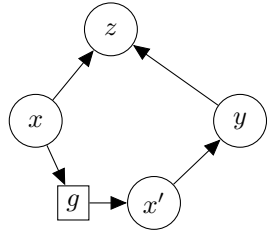
\begin{center}
\vspace{5mm}
\begin{tikzpicture}
  % Define nodes
  \node[latent] (x) {$x$}; % u node
  \node[draw, rectangle, below=0.5cm of x, xshift=0.5cm] (g) {$g$};
  \node[latent, right=0.5cm of g] (gx) {$x'$};
  \node[latent, right=2cm of x] (y) {$y$}; % v node
  \node[latent, above=0.5cm of x, xshift=1cm] (z) {$z$}; % w node
  
  % Connect the nodes
  \edge{x}{g}; % Edge from u to v 
  \edge{g}{gx};
  \edge{gx}{y};
  \edge{x}{z};
  \edge{y}{z};
  % \edge{x}{y};
\end{tikzpicture}
\vspace{5mm}
\end{center}

Effectively, $x'$ works as a mediator between $x$ and $y$. Because $g$ is a deterministic function, the effect of $x$ on $y$ is then perfectly captured by $x'$. In order to understand when this would happen, it helps to consider the structural causal model:\footnote{
    $g$ could take as input noise in addition to $x$, but to strongly emphasize that $x'$ is a nonlinear feature of $x$, we omit it here.
}
\begin{align}
    &x \leftarrow \epsilon_x \\
    &x' \leftarrow g(x) \\
    &y \leftarrow f_y(x, x', \epsilon_y) \\
    &z \leftarrow f_z(x, y, \epsilon_z).
\end{align}

What changes between the last two graphs is the third line in the structural causal model above. The original one is
\begin{align}
    y \leftarrow f_y(x, x', \epsilon_y),
\end{align}
while the new one is
\begin{align}
    y \leftarrow f'_y(x', \epsilon_y).
\end{align}
For this to happen, $x'$ must absorb all relationship between $x$ and $y$. That is, $x'$ must be fully predictive of $y$, leaving only external noise $\epsilon_y$ and nothing more to be captured by $x$. 

Consider a slightly more realistic example of detecting a fox in a picture. There are two major features of any object within any picture; shape and texture. The shape is what we often want our predictor to rely on, while the texture, which is usually dominated by colour information, should be ignored. For instance, if we have a bunch of pictures taken from any place in the sub-arctic Northern Hemisphere, most of the foxes in these pictures will be yellowish with white-coloured breast and dark-coloured feet and tail. 
On the other hand, foxes in the pictures taken in the Arctic will largely be white only, implying that the texture/colour feature of a fox is an environment-dependent feature and is not stable across the environments. Meanwhile, the shape information, a fox-like shape, is the invariant feature of a fox across multiple environments. In this case, $x'$ would be the shape feature of $x$. 

We now see two criteria a function $g$ must satisfy:
\begin{enumerate}
    \item Given $x$ and $y$, $x'=g(x)$ and $z$ are independent.
    \item $x'=g(x)$ is highly predictive of (correlated with) $y$. 
\end{enumerate}
Once we find such $g$, the (potentially biased) outcome can be obtained given a new instance $x$, by fitting a predictive model $\hat{p}(y|x')$~\citep{arjovsky2019invariant}. That is,
\begin{align}
    \hat{y}(x) = \mathbb{E}_{\hat{p}(y|x'=g(x))} \left[ y \right].
\end{align}
This would be free of the spurious correlation arising from the environment condition. 

\paragraph{Learning.}

We now demonstrate one way to learn $g$ to satisfy two conditions above as much as possible. First, in order to satisfy the first condition, we must build a predictor of $z$ given $x'$. This predictor should be non-parametric in order to capture as much (higher-order) correlations that could exist between $z$ and $x'$. Let $\hat{p}(z|x') = h(x')$ be such a predictor obtained by solving the following optimization problem:
\begin{align}
\label{eq:discriminator-training}
    \min_{p} 
    -\frac{1}{N}
    \sum_{n=1}^N
    \log p(z^n | g(x^n)),
\end{align}
where $(x^n, y^n, z^n)$ is the $n$-th training example drawn from the original graph while ensuring that $z^n \in \mathcal{E}$. $\mathcal{E}$ is a set of environments in the training set. In other words, we have a few environments we observe and then condition sampling of $(x,y)$ on, and we use these examples to build an environment predictor from $g(x)$, given $g$. 

The goal is then to minimize the following cost function w.r.t. $g$, where we assume $z$ is discrete:
\begin{align}
    C_1(g) = \sum_{z' \in \mathcal{Z}} \hat{p}(z=z'|x'=g(x)) \log \hat{p}(z=z'|x'=g(x)).
\end{align}
In other words, we maximize the entropy of $\hat{p}(z|x')$, which is maximized when it is uniform. When $\hat{p}(z|x')$ is uniform, it is equivalent to $z \indep x'$. 

One may ask where the condition on observing $y$ went. This is hidden in $\hat{p}(z|x')$, since $\hat{p}$ was estimated using $(g(x),z)$ pairs derived from a set of triples $(x,y,z)$ drawn from the original graph, as clear from Eq.~\eqref{eq:discriminator-training}.  

Of course, this cost function alone is not useful, since it will simply drive $g$ to be a constant function. The second criterion prevents this, and the second criterion can be expressed as
\begin{align}
    C_2(g, q) = 
    -\frac{1}{N}
    \sum_{n=1}^N
    \log q(y^n | g(x^n)).
\end{align}
This second criterion must be minimized with respect to both the feature extractor $g$ and the $y$ predictor $q(y|x')$. This criterion ensures that the feature $x'$ is predictive of (that is, highly correlated with) the output $y$. 

Given $\hat{p}(z|x')$, the feature extractor is then trained to minimize
\begin{align}
\label{eq:feature-training}
    \min_g C_1(g) + \alpha C_2(g,q),
\end{align}
where $\alpha$ is a hyperparameter and balances between $C_1$ and $C_2$. 

We then alterate between solving Eq.~\eqref{eq:discriminator-training} to find $\hat{p}$ and solving Eq.~\eqref{eq:feature-training} to find $g$ and $\hat{q}$~\citep{ganin2016domain}. This is a challenging, bi-level optimization problem and may not even converge both in theory and in practice, although this approach has been used successfully in a few application areas.

The most important assumption here is that we have access to training examples from more than one environments. Preferably, we would have examples from all possible environments (that is, from all possible values $z$ can take), even if they do not necessarily follow $p^*(z|x,y)$ closely. If so, we would simply ignore $z$ by considering $z$ as marginalized. If we have only a small number of environments during training, it will be impossible for us to ensure that $g(x)$ does not encode any information about $z$. There is a connection to generalization, as better generalization in $\hat{p}(z|x')$ would imply a fewer environments necessary for creating a good $\hat{p}(z|x')$ and in turn for producing a more stable feature extractor $g$.

\section{Prediction vs. Causal Inference}

The major difference between prediction and causal inference is the goal. The goal of prediction is to predict which value a particular variable, in our case often the outcome variable, would take given that we have observed the values of the other variables. On the other hand, the goal of causal inference is to know which value the outcome variable would take had we intervened on the action variable. This difference implies that causal inference may not be the best way to predict what would happen based on what we have observed. 

In the example of birds vs. branches above, if our goal is good prediction, we would be certainly open to using the location of the branch as one of the features as well. Even if a large portion of the bird in a picture is occluded by e.g. leaves, we may be able to accurately predict that there is a bird in the picture by noticing the horizontal branch near the bottom of the tree. This branch feature is clearly not a causal feature, but nevertheless helps us make better prediction. In short, if I knew that the picture was taken in a forest, I would rely on both the beak and the branch's location to determine whether there is a bird in the picture. This is however a brittle strategy, as it would certainly degrade my prediction ability had the picture been taken somewhere else.

The invariant predictor $q(y | g(x))$ from above is thus likely sub-optimal in the context of prediction under any environment, although this may be the right distribution to compute the causal effect of $x$ and $y$. This is because the invariant predictor only explains a part of $y$ (marked red below), while ignoring the open path (marked blue below) via the collider:

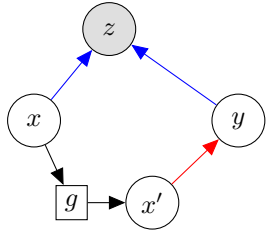
\begin{center}
\vspace{5mm}
\begin{tikzpicture}
  % Define nodes
  \node[latent] (x) {$x$}; % u node
  \node[draw, rectangle, below=0.5cm of x, xshift=0.5cm] (g) {$g$};
  \node[latent, right=0.5cm of g] (gx) {$x'$};
  \node[latent, right=2cm of x] (y) {$y$}; % v node
  \node[obs, above=0.5cm of x, xshift=1cm] (z) {$z$}; % w node
  
  % Connect the nodes
  \edge{x}{g}; % Edge from u to v 
  \edge{g}{gx};
  \edge[color=red]{gx}{y};
  \edge[color=blue]{x}{z};
  \edge[color=blue]{y}{z};
  % \edge{x}{y};
\end{tikzpicture}
\vspace{5mm}
\end{center}

Given an environment $z = \hat{z}$, we must capture both correlations arising from $g(x)\to y$ and $x \to \hat{z} \leftarrow y$, in order to properly predict what value $y$ is likely to take given $x$. This can be addressed by introducing an environment-dependent feature extractor $h_{\hat{z}}(x)$ that is orthogonal to the invariant feature extractor $g(x)$. We can impose such orthogonality (or independence) when learning $h_{\hat{z}}(x)$ by
\begin{align}
    \min_{h, q}
    -\frac{1}{N}
    \sum_{n=1}^N
    \log q(y^n | g(x^n), h_{\hat{z}}(x^n)),
\end{align}
with a given $g$. $h_{\hat{z}}$ would only capture about $y$ that was not already captured by $g$, leading to the orthogonality. This however assumes that $q$ is constrained to the point that it cannot simply ignore $g(x)$ entirely.

This view allows us to use a small number of labelled examples from a new environment in the test time to quickly learn the environment-specific feature extractor $h_z$ while having learned the environment-invariant feature extractor $g$ in the training time from a diverse set of environments. One can view such a scheme as meta-learning or transfer learning, although neither of these concepts is well defined. 

It is possible to flip the process described here to obtain an environment-invariant feature extractor $g$, if we know of an environment-dependent feature extractor $h_z$, by
\begin{align}
    \min_{g, q}
    -\frac{1}{N}
    \sum_{n=1}^N
    \log q(y^n | g(x^n), h_{\hat{z}}(x^n)),
\end{align}
assuming again that $q$ is constrained to the point that it cannot simply ignore $h(x)$ entirely. This flipped approach has been used to build a predictive model that is free of a known societal bias, of which the detector can be easily constructed~\citep{he2019unlearn}.

\section{A Case Study: Language Modeling with Pairwise Preference}

An autoregressive language model is described as a repeated application of the next-token conditional probability, as in
\begin{align}
    p(w_1, w_2, \ldots, w_T) = \prod_{t=1}^T p(w_t | w_{<t}).
\end{align}
A conditional autoregressive language model is exactly the same except that it is conditioned on another variable $X$:
\begin{align}
    p(w_1, w_2, \ldots, w_T | x) = \prod_{t=1}^T p(w_t | w_{<t}, x).
\end{align}
There are many different ways to build a neural network to implement the next-token conditional distribution. We do not discuss any of those approaches, as they are out of the course's scope. 

An interesting property of a language model is that it can be used for two purposes:
\begin{enumerate}
    \item Scoring a sequence: we can use $p(w_1, w_2, \ldots, w_T | X)$ to score an answer sequence $w$ given a query $x$.
    \item Approximately finding the best sequence: we can use approximate decoding to find $\arg\max_w p(w | x)$. 
\end{enumerate}
This allows us to perform causal inference and outcome maximization simultaneously.

Consider the problem of query-based text generation, where the goal is to produce an open-ended answer $w$ to a query $x$. Because it is often impossible to give an absolute score to the answer $w$ given a query $x$, it is customary to ask a human annotator a relative ranking between two (or more) answers $w_+$ and $w_-$ given a query $x$. Without loss of generality, let $w_+$ be the preferred answer to $w_-$. 

We assume that there exists a strict total order among all possible answers. That is,
\begin{enumerate}
    \item Irreflexive: $r(w|x) < r(w|x)$ cannot hold.
    \item Asymmetric: If $r(w|x) < r(w'|x)$, then $r(w|x) > r(w'|x)$ cannot hold.
    \item Transitive: If $r(w|x) < r(w'|x)$ and $r(w'|x) < r(w''|x)$, then $r(w|x) < r(w''|x)$.
    \item Connected: If $w \neq w'$, then either $r(w|x) < r(w'|x)$ or $r(w|x) > r(w'|x)$ holds.
\end{enumerate}
In other words, we can enumerate all possible answers according to their (unobserved) ratings on a 1-dimensional line.

\paragraph{A non-causal approach.}

It is then relatively trivial to train this language model, assuming that we have a large amount of triplets 
\[
D=\left\{(x^1, w^1_+, w^1_-), \ldots, (x^N, w^N_+, w^N_-)\right\}.
\]
For each triplet, we ensure that the language model puts a higher probability on $w_+$ than on $w_-$ given $x$ by minimizing the following loss function:
\begin{align}
    L_{\mathrm{pairwise}}(p) =
    \frac{1}{N}
    \sum_{n=1}^N
    \max(0, m-\log p(w^n_+|x) + \log p(w^n_- | x)),
\end{align}
where $m \in [0, \infty)$ is a margin hyperparameter. For each triplet, the loss inside the summation is zero, if the language model puts the log-probability on $w_+$ more than that on $w_-$ with the minimum margin of $m$. 

This loss alone is however not enough to train a well-trained language model from which we can produce a high-quality answer. For we have only pair-wise preference triplets for reasonable answers only. The language model trained in this way is not encouraged to put low probabilities on gibberish. We avoid this issue by ensuring that the language model puts reasonably high probabilities on all reasonable answer by minimizing the following extra loss function:
\begin{align}
    L_{\mathrm{likelihood}}(p) =
    -
    \frac{1}{2N}
    \sum_{n=1}^N
    \left(
    \log p(w^n_+ | x)
    +
    \log p(w^n_- | x)
    \right),
\end{align}
which corresponds to the so-called negative log-likelihood loss. 

\paragraph{A causal consideration.}

This approach works well under the assumption that it is only the content that is embedded in the answer $w$. This is unfortunately not the case. Any answer is a combination of the content and the style, and the latter should not be the basis on which the answer is rated. For instance, one aspect of style is the verbosity. Often, a longer answer is considered to be highly rated, because of the subconscious bias by a human rater believing a better answer would be able to write a longer answer, although there is no reason why there should not be a better and more concise answer. 

This process can be described as the graph below, where $r$ is the rating and $s$ is the style:

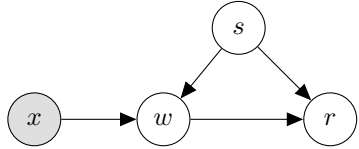
\begin{center}
\vspace{5mm}
\begin{tikzpicture}
  % Define nodes
  \node[latent] (w) {$w$}; 
  \node[latent, right=1.5cm of w] (r) {$r$}; 
  \node[latent, above=0.5cm of w, xshift=1cm] (s) {$s$}; 
  \node[obs, left=1cm of w] (x) {$x$}; 
  
  % Connect the nodes
  \edge{w}{r}; % Edge from u to v 
  \edge{s}{w};
  \edge{s}{r};
  \edge{x}{w};
  % \edge{x}{r};
\end{tikzpicture}
\vspace{5mm}
\end{center}

The direct effect of $w$ on the rating $r$ is based on the content, but then there is spourious correlation between $w$ and $r$ via the style $s$. For instance, $s$ could encode the verbosity which affects both how $w$ is written and how a human rater perceives the quality and gives the rating $r$. In the naive approach above, the language model, as a scorer, will fail to distinguish between these two and capture both, which is clearly undesirable; a longer answer is not necessarily a better answer. In other words, a language model $p_0$ trained in a purely supervised learning way above will score $w$ high for both causal and spurious (via $s$) reasons. An answer $w$ sampled from $p_0$ can then be considered dependent upon not only the question $x$ itself but also of an unobserved style variable $s$. 

\paragraph{Direct preference optimization~\citep{rafailov2024direct} or unlikelihood learning~\citep{welleck2019neural}.}

We can resolve this issue by combining two ideas we have studied earlier; randomized controlled trials (RCT; \S\ref{sec:randomized-controlled-trials}) and inverse probability weighting (IPW; \S\ref{sec:ipw}). First, we sample two answers, $w$ and $w'$, from the already trained model $p_0$, using supervised learning above:
\begin{align}
    w, w' \sim p_0(w|x).
\end{align}
These two answers (approximately) maximize the estimated outcome (rating) by capturing both the content and style. 
One interesting side-effect of imperfect learning and inference (generation) is that both of these answers would largely share the style. If we use $s'$ to denote that style, we can think of each answer as sampled from $w | x, s'$. 
With a new language model $p_1$ (potentially initialized from $p_0$), we can compute the rating after removing the dependence on the style $s$ by IPW:
\begin{align}
    \hat{r}(w|x) = \frac{p_1(w|x)}{p_0(w|x)}. 
\end{align}
This reminds us of $\mathrm{do}$ operation, resulting in the following modified graph:
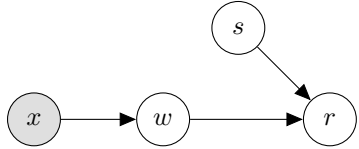
\begin{center}
\vspace{5mm}
\begin{tikzpicture}
  % Define nodes
  \node[latent] (w) {$w$}; 
  \node[latent, right=1.5cm of w] (r) {$r$}; 
  \node[latent, above=0.5cm of w, xshift=1cm] (s) {$s$}; 
  \node[obs, left=1cm of w] (x) {$x$}; 
  
  % Connect the nodes
  \edge{w}{r}; % Edge from u to v 
  \edge{s}{r};
  \edge{x}{w};
  % \edge{x}{r};
\end{tikzpicture}
\vspace{5mm}
\end{center}

Of course, this score $\hat{r}$ does not mean anything, since $p_1$ does not mean anything yet. We have to train $p_1$ by asking an expert to provide their preference between $w$ and $w'$. Without loss of generality, let $w$ be the preferred answer over $w'$. That is, $w_+=w$ and $w_-=w'$. We train $p_1$ by minimizing
\begin{align}
    L'_{\mathrm{pairwise}}(p_1) =
    \frac{1}{N}
    \sum_{n=1}^N
    \max\left(0, m-\log \frac{p_1(w^n_+|x)}{p_0(w^n_+|x)} + 
    \log \frac{p_1(w^n_- | x)}{p_0(w^n_- | x)}
    \right),
\end{align}
where we assume have $N$ pairs. $m$ is a margin as before. It is possible to replace the margin loss with another loss function, such as a log loss or linear loss.

This procedure encourages $p_1$ to capture only the direct (causal) effect of the answer on the rating, dissecting out the indirect (spurious) effect via the style $s$. One training is done, we use $p_1$ to produce a better answer, which dependes less on the spurious correlation between the answer and the rating via the style. 

Because this procedure is extremely implicit about the existence of and the dependence on the style, it can be beneficial to repeat this procedure multiple rounds in order to further remove the effect of the spurious correlation and improve the quality of a generated answer~\citep{ouyang2022training}. 

\section{Summary}

In this chapter, we have learned the following concepts:

\begin{enumerate}
    \item Out-of-distribution generalization and its impossibility
    \item Invariance as a core principle behind out-of-distribution generalization
    \item Preference modeling for training a language model, as causal learning
\end{enumerate}

The goal of this chapter has been to introduce students to the concept of learning beyond independently-and-identically-distribution settings, by relying on concepts and frameworks from causal inference and more broadly causality. The topics covered in this chapter are sometimes referred to as {\it causal machine learning}~\citep{kaddour2022causal}.

\chapter{Remaining Topics}
\label{chap:remaining-topics}

As the purpose of this course is to be a thin and quick introductory course at the intersection of causal inference and machine learning, it is not the intention nor desirable to cover all topics in causal inference exhaustively. In this final chapter, I discuss a few topics that I did not feel necessary to be included in the main course but could be useful for students if they could be taught.

\section{Other Techniques in Causal Inference}

In practice the following observational causal inference techniques are widely used:
\begin{itemize}
    \item Regression in \S\ref{sec:regression}
    \item Inverse probability weighting in \S\ref{sec:ipw} and Matching in \S\ref{sec:matching}
    \item Instrument variables in \S\ref{sec:instrumental-variables}
    \item Difference-in-difference
    \item Regression discontinuity design
    \item Double machine learning
\end{itemize}

Difference-in-difference and regression discontinuity design are heavily used in practice, but they work for relatively more specialized cases, which is why this course has omitted them so far. In this section, we briefly cover these two approaches for the sake of completeness. Furthermore, this section wraps up by providing a high-level intuition behind a more recently proposed and popularized technique of double machine learning.

\subsection{Difference-in-Difference}

The average treatment effect (ATE) from \S\ref{sec:ate} measures the difference between the outcomes of two groups; treated and not treated, or more precisely, it measures the difference between the outcome of the treated group and the expected outcome over all possible actions. 

One way to interpret this is to view ATE as checking what happens to a treated individual had the individual was not treated, on average. First, we can compute what happens to the individual once they were treated, on average, as
\begin{align}
    y^1_{\mathrm{diff}} = \mathbb{E}_x \mathbb{E}_a \mathbb{E}_{y_{\mathrm{pre}},y_{\mathrm{post}}} \left[ \mathds{1}(a = 1) (y_{\mathrm{post}} - y_{\mathrm{pre}}) \right],
\end{align}
where $y_{\mathrm{pre}}$ and $y_{\mathrm{post}}$ are the outcomes before and after the treatment ($a=1$). We can similarly compute what happens to the individual had they not been treated, on average, as well by
\begin{align}
    y^0_{\mathrm{diff}} = \mathbb{E}_x \mathbb{E}_a \mathbb{E}_{y_{\mathrm{pre}},y_{\mathrm{post}}} \left[ \mathds{1}(a = 0) (y_{\mathrm{post}} - y_{\mathrm{pre}}) \right].
\end{align}

We now check the difference between these two quantities:
\begin{align}
\label{eq:diff-in-diff}
    y^1_{\mathrm{diff}} - y^0_{\mathrm{diff}}
    =
    \mathbb{E}_x \mathbb{E}_a
    \left[
    \right.
    &
    \mathbb{E}_{y_{\mathrm{post}}} \left[
    \mathds{1}(a = 1) y_{\mathrm{post}}
    -
    \mathds{1}(a = 0) y_{\mathrm{post}}
    \right]
    \\
    -
    &
    \left.
    \mathbb{E}_{y_{\mathrm{pre}}} \left[
    \mathds{1}(a = 1) y_{\mathrm{pre}}
    -
    \mathds{1}(a = 0) y_{\mathrm{pre}}
    \right]
    \right].
\end{align}

If we used RCT from \S\ref{sec:randomized-controlled-trials} to assign the action independent of the covariate $x$ and also {\it uniformly}, the second term, that is the difference in the pre-treatment outcome, should disappear, since the treatment had not been given to the treatment group yet. This leaves only the first term, which is precisely how we would compute the outcome from RCT. 

In an observational study, that is passive causal inference, we often do not have a control over how the participants were split into treatment and placebo groups. This often leads to the discrepancy in the base outcome between the treated and placebo groups. In that case, the second term above would not vanish but will work to remove this baseline effect. 

Consider measuring the effect of a vitamin supplement on the height of school-attending girls of age 10. Let us assume that this particular vitamin supplement is provided to school children by default in Netherlands from age 10 but is not in North Korea. We may be tempted to simply measure the average heights of school-attending girls of age 10 from these two countries, and draw a conclusion whether this supplement helps school children grow taller. This however would not be a reasonable way to draw the conclusion, since the averages heights of girls of age 9, right before the vitamin supplement begins to be provided in Netherlands, differ quite significantly between two countries (146.55cm vs. 140.58cm.) We would rather look at how much taller these children grew between ages of 9 and 10. 

Because we consider the difference of the difference in Eq.~\eqref{eq:diff-in-diff}, we call this estimator {\it difference-in-difference}. This approach is widely used and was one of the most successful cases of passive causal inference, dating back to the 19th century~\citep{snow1856mode}.

In the context of what we have learned this course, let us write a structural causal model that admits this difference-in-different estimator:
\begin{align}
    &x \leftarrow \epsilon_x \\
    &a \leftarrow \mathds{1}(x + \epsilon_a) \\
    &y \leftarrow \mathds{1}(x > 0) y_0 + \alpha a + \epsilon_y.
\end{align}
With zero-mean and symmetric $\epsilon_x$ and $\epsilon_a$, those with positive $x$ are more likely to be assigned to $a=1$. Due to the first term in $y$, the outcome has a constant bias $y_0$ when $x$ is positive. In other words, those, who are likely to be given the treatment, have $y_0$ added to the outcome regardless of the treatment ($a=1$) itself, since $+y_0$ does not depend on $a$. The difference-in-difference estimator removes the effect of $y_0$ from estimating $\alpha$ which is the direct causal effect of $a$ on $y$. 

This tells us when the difference-in-difference estimator works, and how we can extend it further. For instance, it is not necessary to assume the linearity between $a$ and $y$. I leave it to you as an exercise.

\subsection{Regression Discontinuity}

Another popular technique for passive causal inference is called regression discontinuity~\citep[see][and references therein]{imbens2008regression}. Regression discontinuity assumes that there exists a simple rule to determine to which group, either treated or placebo, an individual is assigned based on the covariate $x$. This rule can be written down as
\begin{align}
\label{eq:rd-rule}
    a = 
    \begin{cases}
        1, &\text{ if } x_d \geq c_0 \\
        0, &\text{ otherwise}.
    \end{cases}
\end{align}
If the $d$-th covariate crosses over the threshold $c_0$, the individual is assigned to $a=1$. 

We further assume that the outcome given a particular action is a smooth function of the covariate. That is, the outcome of a particular action, $f(\hat{a}, x)$, changes smoothly especially around the threshold $c_0$. In other words, had it not been for the assignment rule above, $\lim_{x_d \to c_0} f(\hat{a}, x) = \lim_{c_0 \leftarrow x_d} f(\hat{a}, x)$. There is no discontinuity of $f(\hat{a}, x)$ at $x_d=c_0$, and we can fit a smooth predictor that extrapolates well to approximate $f(\hat{a}, x)$ (or $\mathbb{E}_{x_{d' \neq d}} f(\hat{a}, x_{d'}\cup x_c=c_0)$.) 

If we assume that the threshold $c_0$ was chosen arbitrarily, that is independent of the values of $x_{\neq d}$, it follows that the distributions over $x_{\neq d}$ before and after $c_0$ to remain the same at least locally.\footnote{
    This provides a good ground for testing the validity of regression discontinuity. If the distributions of $x$ before and after $c_0$ differ significantly from each other, regression discontinuity cannot be used. 
}
This means that the assignment of an action $a$ and the covariate other than $x_d$ are independent locally, i.e., $|x_d - c_0| \leq \epsilon$, where $\epsilon$ defines the radius of the local neighbourhood centered on $c_0$. Thanks to this independence, which is the key difference between the conditional and interventional distributions, as we have seen repeatedly earlier, we can now compute the average treatment effect locally (so is often called a {\it local average treatment effect}) as
\begin{align}
\mathrm{LATE} 
=&
    \mathbb{E}_{x}
    \left[ 
    \mathds{1}(|x_d - c_0| \leq \epsilon)
    f(1, x)
    -
    f(0, x)
    \right]
    \\
    =&
    \mathbb{E}_{x: |x_d - c_0| \leq \epsilon}
    \left[
    f(1,x)
    \right]
    -
    \mathbb{E}_{x: |x_d - c_0| \leq \epsilon}
    \left[
    f(0,x)
    \right].
\end{align}

Of course, our assumption here is that we do not observed $x_{\neq d}$. Even worse, we never observe $f(1,x)$ when $x_d < c_0$ and $f(0,x)$ when $x_d > c_0$. Instead, we can fit a non-parametric regression model $\hat{f}(\hat{a}, x_d)$ to approximate $\mathbb{E}_{x_{\neq d} | x_d} f(\hat{a},x)$ and expect (or hope?) that it would extrapolate either before or after the threshold $c_0$. Then, LATE becomes
\begin{align}
\mathrm{LATE} 
&=
\int_{c_0-\epsilon}^{c_0+\epsilon}
    \hat{f}(1,x_d)
    -
    \hat{f}(0,x_d)
    \mathrm{d}x_d
    \\
&=_{\epsilon \to 0}
\hat{f}(1, c_0)
-
\hat{f}(0, c_0),
\end{align}
thanks to the smoothness assumption of $f$. 

The final line above tells us pretty plainly why this approach is called regression discontinuity design. We literally fit two regression models on the treated and placebo groups and look at their discrepancy at the decision threshold. The amount of the discrepancy implies the change in the outcome due to the change in the action, of course under the strong set of assumptions we have discussed so far.

\subsection{Double Machine Learning}

Recent advances in machine learning have open a door to training large-scale non-parametric methods on high-dimensional data. This allows us to expand some of the more conventional approaches. One such example is double machine learning~\citep{chernozhukov2018double}. We briefly describe one particular instantiation of double machine learning here. 

Recall the instrument variable approach from \S\ref{sec:instrumental-variables}. The basic idea was to notice that the action $a$ was determined using two independent sources of information, the confounder $x$ and the external noise $\epsilon_a$:
\begin{align}
    a \leftarrow f_a(x, \epsilon_a),
\end{align}
with $x \indep \epsilon_a$. We then introduced an instrument $z$ that is a subset of $\epsilon_a$, such that $z$ is predictive of $a$ but continues to be independent of $x$. From $z$, using regression, we capture a part of variation in $a$ that is independent of $x$, in order to severe the edge from the confounder $x$ to the outcome $y$. Then, we use this instrument-predicted action $a'$ to predict the outcome $y$. 
We can instead think of fitting a regression model $g_a$ from $x$ to $a$ and use the residual $a_{\bot} = a - g_a(x)$ as the component of $a$ that is independent of $x$, because the residual was not predictable from $x$. 

This procedure can now be applied to the outcome which is written down as
\begin{align}
    y \leftarrow f_y(a_{\bot}, x, \epsilon_y).
\end{align}
Because $x$ and $a_{\bot}$ are independent, we can estimate the portion of $y$ that is predictable from $y$ by building a predictor $g_y$ of $y$ given $x$. The residual $y_{\bot} = y - g_y(x)$ is then what cannot be predicted by $x$, directly nor via $a$. We are in fact relying on the fact that such a non-parametric predictor would capture both causal and spurious correlations indiscriminately.

$a_{\bot}$ is a subset of $a$ that is independent of the confounder $x$, and $y_{\bot}$ is a subset of $y$ that is independent of the confounder $x$. The relationship between $a_{\bot}$ and $y_{\bot}$ must then be the direct causal effect of the action on the outcome. In other words, we have removed the effect of $x$ on $a$ to close the backdoor path, resulting in $a_{\bot}$. We have removed the effect of $x$ on $y$ to reduce non-causal noise, resulting $y_{\bot}$. What remains is the direct effect of $a$ on the outcome $y$. We therefore fit another regression from $a_{\bot}$ to $y_{\bot}$, in order to capture this remaining correlation that is equivalent to the direct cause of $a$ on $y$. 

\section{Behaviour Cloning from Multiple Expert Policies Requires a World Model}

A Markov decision process (MDP) is often described as a tuple of the following items:
\begin{enumerate}
    \item $\mathcal{S}$: a set of all possible states
    \item $\mathcal{A}$: a set of all possible actions
    \item $\tau: \mathcal{S} \times \mathcal{A} \times \mathcal{E} \to \mathcal{S}$: a transition dynamics. $s' = \tau(s, a, \epsilon)$.
    \item $\rho: \mathcal{S} \times \mathcal{A} \times \mathcal{S} \to \mathbb{R}$: a reward function. $r = \rho(s, a, s')$.
\end{enumerate}

The transition dynamics $\tau$ is a deterministic function but takes as input noise $\epsilon \in \mathcal{E}$, which overall makes it stochastic. We use $p_\tau(s' | s, a)$ to denote the conditional distribution over the next state given the current state and action by marginalizing out noise $\epsilon$. The reward function $r$ depends on the current state, the action taken and the next state. It is however often the case that the reward function only depends on the next (resulting) state. 

A major goal is then to find a policy $p_\pi: \mathcal{S} \times \mathcal{A} \to \mathbb{R}_{>0}$ that maximizes
\begin{align}
\label{eq:return}
J(\pi) =& 
    \sum_{s_0} p_0(s_0) 
    \sum_{a_0} p_\pi(s_0, a_0) 
    \sum_{s_1} p_{\tau}(s_1 | s_0, a_0) 
    \left(\gamma^0 \rho(s_0, a_0, s_1)
    \right.
    \nonumber
    \\
    &
    \qquad
    \quad
    +
    \left.
    \sum_{a_1} p_\pi(s_1, a_1) \sum_{s_2} p_{\tau}(s_2 | s_1, a_1)
    \left(
    \gamma^1 \rho(s_1, a_1, s_2)
    +
    \cdots
    \right)
    \right) 
    \\
    =&
    \mathbb{E}_{s_0 \sim p_0(s_0)}
    \mathbb{E}_{a_0, s_1 \sim p_\pi(a_0|s_0) p_{\tau}(s_1|s_0,a_0)}
    \nonumber
    \\
    &
    \qquad
    \qquad
    \quad
    \mathbb{E}_{a_1, s_2 \sim p_\pi(a_1|s_1) p_{\tau}(s_2|s_1,a_1)}
    \cdots
    \left[
    \sum_{t=0}^\infty \gamma^t \rho(s_t, a_t, s_{t+1})
    \right]
    \\
    =&
    \mathbb{E}_{p_0, p_\pi, p_\tau}
    \left[ 
    \sum_{t=0}^\infty \gamma^t \rho(s_t, a_t, s_{t+1})
    \right]
    ,
\end{align}
where $p_0(s_0)$ is the distribution over the initial state. 

$\gamma \in (0, 1]$ is a discounting factor. The discounting factor can be viewed from two angles. First, we can view it conceptually as a way to express how much we care about the future rewards. With a large $\gamma$, our policy can sacrifice earlier time steps' rewards in return of higher rewards in the future. The other way to think of the discounting factor is purely computational. With $\gamma < 1$, we can prevent the total return $J(\pi)$ from diverging to infinity, even when the length of each episode is not bounded.

As we have learned earlier when we saw the equivalence between the probabilistic graphical model and the structural causal model in \S\ref{sec:pgm}--\ref{sec:scm}, we can guess the form of $\pi$ as a deterministic function:
\begin{align}
    a \leftarrow \pi(s, \epsilon_\pi).
\end{align}
Together with the transition dynamics $\tau$ and the reward function $\rho$, we notice that the Markov decision process can be thought of as defining a structural causal model for each time step $t$ as follows:
\begin{align}
    &s\text{ is given.} \\
    &a \leftarrow \pi(s, \epsilon_\pi) \\
    &s' \leftarrow \tau(s, a, \epsilon_{s'}) \\
    &r \leftarrow \rho(s', \epsilon_{r}),
\end{align}
where we make a simplifying assumption that the reward only depends on the landing state.

Graphically,
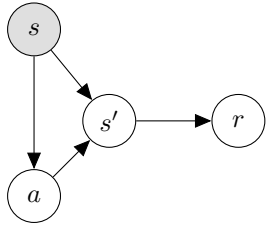
\begin{center}
\vspace{5mm}
\begin{tikzpicture}
  % Define nodes
  \node[obs] (s) {$s$}; 
  \node[latent, below=1.5cm of s] (a) {$a$}; 
  \node[latent, below=0.5cm of s, xshift=1cm] (sn) {$s'$}; 
  \node[latent, right=1cm of sn] (r) {$r$}; 
  
  % Connect the nodes
  \edge{s}{a}; % Edge from u to v 
  \edge{s}{sn};
  \edge{a}{sn};
  \edge{sn}{r};
  % \edge{x}{r};
\end{tikzpicture}
\vspace{5mm}
\end{center}

\paragraph{Behaviour cloning.}

With this in our mind, let us consider the problem of so-called `behavior cloning'. In behaviour cloning, we assume the existence of an expert policy $\pi^*$ that results in a high return $J(\pi^*)$ from Eq.~\eqref{eq:return} and that we have access to a large amount of data collected from the expert policy. This dataset consists of tuples of current state $s$, action by the expert policy $a$ and the next state $s'$. We often do not observe the associated reward directly.
\begin{align}
    D = \left\{ (s_n, a_n, s'_n) \right\}_{n=1}^N,
\end{align}
where $a_n \sim p_{\pi^*}(a | s_n)$ and $s'_n \sim p_{\tau}(s' | s_n, a_n)$.

Behavior cloning refers to training a policy $\pi$ that imitates the expert policy $\pi^*$ using this dataset. We train a new policy $\pi$ often by maximizing
\begin{align}
\label{eq:bc-loss}
    J_{\mathrm{bc}}(\pi) = \sum_{n=1}^N \log \pi(a_n, s_n).
\end{align}
In other words, we ensure that the learned policy $\pi$ puts a high probability on the action that was taken by the expert policy $\pi^*$. 

\paragraph{Behaviour cloning with multiple experts.}

It is however often that it is not just one expert policy that was used to collect data but a set of expert-like policies that collected these data points.
It is furthermore often that we do not know which such expert-like policy was used to produce each tuple $(s_n, a_n, s'_n)$. This necessitates us to consider the policy used to collect these tuples as a random variable that we do not observe, resulting the following graphical model:\footnote{
    I am only drawing two time steps for simplicity, however, without loss of generality.
}
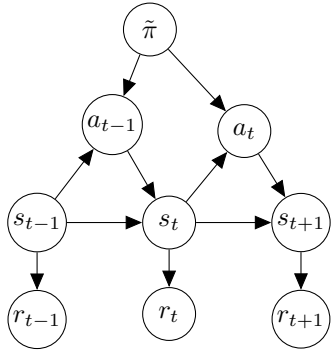
\begin{center}
\vspace{5mm}
\begin{tikzpicture}
  % Define nodes
  \node[latent] (sm1) {$s_{t-1}$};
  \node[latent, right=1cm of sm1] (s) {$s_t$};
  \node[latent, right=1cm of s] (sp1) {$s_{t+1}$};

  \node[latent, below=0.5cm of sm1] (rm1) {$r_{t-1}$};
  \node[latent, below=0.5cm of s] (r) {$r_{t}$};
  \node[latent, below=0.5cm of sp1] (rp1) {$r_{t+1}$};

  \node[latent, above=0.5cm of sm1, xshift=1cm] (am1) {$a_{t-1}$}; 
  \node[latent, above=0.5cm of s, xshift=1cm] (a) {$a_{t}$}; 

  \node[latent, above=0.5cm of am1, xshift=0.5cm] (pi) {$\tilde{\pi}$}; 
  
  % Connect the nodes
  \edge{sm1}{s}; % Edge from u to v 
  \edge{s}{sp1};
  \edge{am1}{s}; % Edge from u to v 
  \edge{a}{sp1};
  \edge{sm1}{am1};
  \edge{s}{a};
  \edge{pi}{am1};
  \edge{pi}{a}; 

  \edge{sm1}{rm1};
  \edge{s}{r};
  \edge{sp1}{rp1};
  % \edge{x}{r};
\end{tikzpicture}
\vspace{5mm}
\end{center}

The inclusion of an unobserved $\tilde{\pi}$ makes the original behaviour cloning objective in Eq.~\eqref{eq:bc-loss} less than ideal. In the original graph, because we sampled both $s$ and $a$ without conditioning on $s'$, there was only one open path between $s$ and $a$, that is, $s\to a$. We could thereby simply train a policy to capture the correlation between $s$ and $a$ to learn the policy which should capture $p(a | \mathrm{do}(s))$. With the unobserved variable $\pi$, this does not hold anymore. 

Consider $(s_t,a_t)$. There are two open paths between these two variables. The first one is the original direct path; $s_t \to a_t$. There is however the second path now; $s_t \leftarrow a_{t-1} \leftarrow \pi \rightarrow a_t$. If we na\"ively train a policy $\pi$ on this dataset, this policy would learn to capture the correlation between the current state and associated action arising from both of these paths. This is not desirable as the second path is {\it not} causal, as we discussed earlier in \S\ref{sec:confounders-colliders-mediators}. In other words, $\pi(a | s)$ would not correspond to $p(a | \mathrm{do}(s))$. 

In order to block this backdoor path, we can use the idea of inverse probability weighting (IPW; \S\ref{sec:ipw}). If we assume we have access to the transition model $\tau$, we can use it to severe two direct connections into $s_t$; $s_{t-1} \to s_t$ and $a_{t-1} \to s_t$, by
\begin{align}
\mathbb{E}_{a_t \sim p_{\pi}(a_t | \mathrm{do}(s_t))}[a_t]
=
\mathbb{E}_{s_t}
\left[
    \frac{p_{\pi}(a_t | s_t)}{
    p_{\tau}(s_t | s_{t-1}, a_{t-1})
    }
    a_t
\right].
\end{align}

\paragraph{Learned transition: a world model.}

Of course, we often do not have access to $\tau$ directly, but must infer this transition dynamics from data. 
Unlike the policy $s \to a$, fortunately, the transition $(s, a) \to s'$ is however not confounded by $\pi$. We can therefore learn an approximate transition model, which is sometimes referred to as a world model~\citep[][and references therein]{lecun2022path}, from data. 
This can be done by 
\begin{align}
    \hat{\tau} = \arg\max_{\tau} \sum_{n=1}^N \log p_{\tau} (s'_n | s_n, a_n).
\end{align}

\paragraph{Deconfounded behaviour closing.}

Once training is done, we can use $\hat{\tau}$ in place of the true transition dynamics $\tau$, to train a de-confounded policy by
\begin{align}
    \hat{\pi} = \arg\max_{\pi} \sum_{n=1}^N \log 
    \frac{p_{\pi}(a'_n | s'_n)}
    {p_{\hat{\tau}}(s'_n | s_n, a_n)},
\end{align}
where $a'_n$ is the next action in the dataset. That is, the dataset now consists of $(s_n, a_n, s'_n, a'_n)$ rather than $(s_n, a_n, s'_n)$. This effectively makes us lose a few examples from the original dataset that correspond to the final steps of episodes, although this is a small price to pay to avoid the confounding by multiple expert policies. 

\paragraph{Causal reinforcement learning.}
This is an example of how causality can assist us in identifying a potential issue {\it a priori} and design a better learning algorithm without relying on trials and errors. In the context of reinforcement learning, which is a sub-field of machine learning focused on learning a policy, such as like behaviour cloning, this is often referred to as and studied in causal reinforcement learning~\citep{bareinboim2020}.

\section{Summary}

In this final chapter, I have touched upon a few topics that were left out from the main chapters perhaps for no particular strong reason. These topics included
\begin{enumerate}
    \item Difference-in-Difference
    \item Regression discontinuity
    \item Double machine learning
    \item A taste of causal reinforcement learning
\end{enumerate}

There are many interesting topics that were not discussed in this lecture note both due to the lack of time as well as the lack of my own knowledge and expertise. I find the following two areas to be particular interesting and recommend you to follow up on.
\begin{enumerate}
    \item Counterfactual analysis: Can we build an algorithm that can imagine taking an alternative action and guess the resulting outcome instead of the actual outcome?
    \item (Scalable) causal discovery: How can we infer useful causal relationship among many variables? 
    \item Beyond invariance (\S\ref{sec:invariance}): Invariance is a strong assumption. Can we relax this assumption to identify a more flexible notion of causal prediction?
\end{enumerate}

\bibliography{main}
\bibliographystyle{abbrvnat}

\end{document}